\documentclass[journal]{IEEEtran}
\usepackage{amsmath}
\usepackage{bm}
\usepackage{hyperref}
\usepackage{url}
\usepackage{color}
\usepackage{siunitx}
\usepackage{graphicx}
\usepackage{makecell}

\usepackage{amsfonts}
\usepackage{amssymb}
\usepackage[mathscr]{euscript}
\usepackage{comment}
\usepackage{subfigure}
\usepackage{algorithm}
\usepackage{algorithmic}
\usepackage{multirow}
\usepackage{wrapfig}
\usepackage{booktabs}
\usepackage{threeparttable}
\usepackage[english]{babel}
\usepackage{microtype}
\usepackage{xcolor}

\newcommand{\etal}{\textit{et al.}}
\DeclareMathOperator*{\argmax}{arg\,max}

\newcommand{\para}[1]{\noindent\textbf{#1}}

\begin{document}

\title{Prior-Guided Multi-View 3D Head Reconstruction}

\author{Xueying~Wang, Yudong~Guo, Zhongqi~Yang and Juyong~Zhang$^\dagger$
\thanks{The authors are with the School of Mathematical Sciences, University of Science and Technology of China.}
\thanks{$^\dagger$Corresponding author. Email: {\texttt{juyong@ustc.edu.cn}}.}}

% \markboth{IEEE Transactions on Multimedia}%
% {Wang \MakeLowercase{\textit{et al.}}: Prior-Guided Multi-View Neural 3D Head Reconstruction}

\maketitle

\begin{abstract}
Recovery of a 3D head model including the complete face and hair regions is still a challenging problem in computer vision and graphics. In this paper, we consider this problem using only a few multi-view portrait images as input. Previous multi-view stereo methods that have been based, either on optimization strategies or deep learning techniques, suffer from low-frequency geometric structures such as unclear head structures and inaccurate reconstruction in hair regions. To tackle this problem, we propose a prior-guided implicit neural rendering network. Specifically, we model the head geometry with a learnable signed distance field (SDF) and optimize it via an implicit differentiable renderer with the guidance of some human head priors, including the facial prior knowledge, head semantic segmentation information and 2D hair orientation maps. The utilization of these priors can improve the reconstruction accuracy and robustness, leading to a high-quality integrated 3D head model. Extensive ablation studies and comparisons with state-of-the-art methods demonstrate that our method can generate high-fidelity 3D head geometries with the guidance of these priors.

\end{abstract}


\IEEEpeerreviewmaketitle

\begin{figure*}[t]
	\centering	\includegraphics[width=1.0\textwidth]{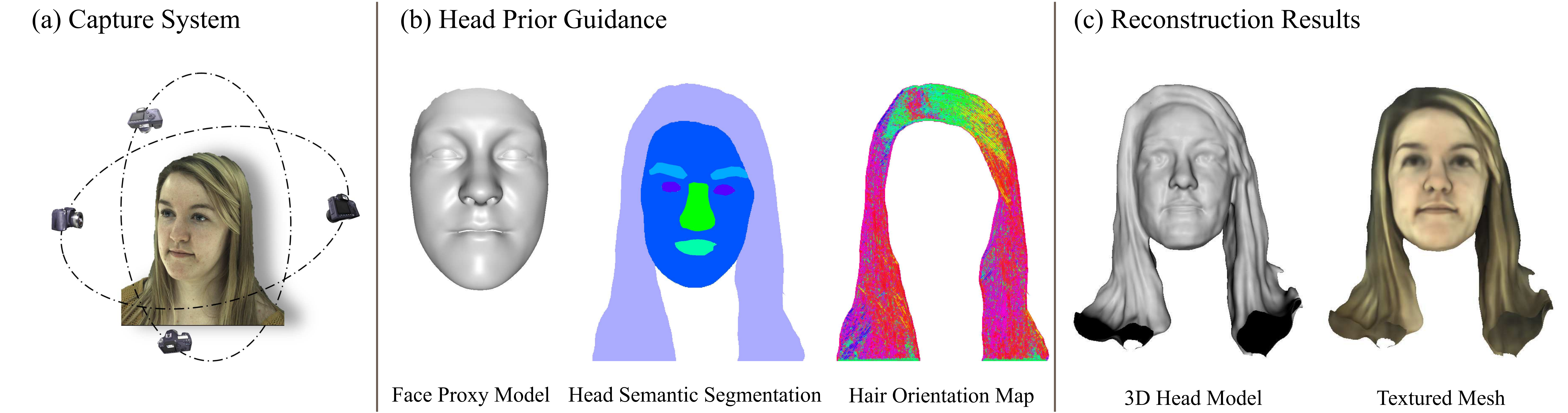}
	\caption{In this paper, we present a novel prior-guided implicit neural rendering method for multi-view stereo-based 3D head reconstruction. Given a few captured multi-view portrait images (a),  our method can obtain three different head priors, including facial prior knowledge, head semantic segmentation information and hair orientation maps (here we show a semantic mask and a hair orientation map under the front view) (b). With the guidance of this prior knowledge, we can recover the high-quality 3D head model (c) via an implicit differentiable renderer.}
	\label{fig:teaser}
\end{figure*}

\section{Introduction}

High-fidelity 3D head reconstruction is an important topic in computer vision and graphics with various applications such as game character generation \cite{game1,game2}, film production \cite{song2020accurate,zhou20183d} and virtual reality \cite{anbarjafari20173d,lifkooee2018image,DBLP:journals/tmm/LouWNHMWY20,wenninger2020realistic}. While head modeling in the real world can be directly performed by scanning the human head, this approach requires the use of expensive equipment such as laser scanners, and is not suitable for consumer-level production. To reduce the setup costs, single image-based reconstruction methods have recently attracted increasing attention, such as optimization-based methods \cite{jiang20183d,tran2018extreme}, deep learning-based methods \cite{guo2018cnn,chen2019photo,wang2020lightweight,tu20203d,fan2020dual} and the dictionary learning-based method \cite{7932891}. Although most of these methods can reconstruct quite good face models, their reconstructed models usually do not include the hair region due to the limited information from the single image. To enable complete head model reconstruction, we focus on multi-view stereo (MVS) reconstruction methods and aim to reconstruct a full head model including the hair region from a few multi-view images.

Multi-view stereo reconstruction \cite{furukawa2015multi} is a commonly used approach for the recovery of complete 3D shapes from multi-view images with overlapping areas. Unfortunately, traditional optimization-based MVS methods mainly suffer from the requirements for a large number of input images and high computational cost. Moreover, their output results are often noisy and incomplete with regard to texture-less, specular and reflective regions due to the inaccurate matching of hand-crafted features~\cite{galliani2015massively, schonberger2016pixelwise, wu2016fast}. Recently, deep learning-based multi-view stereo reconstruction methods have achieved great progress in overcoming the limitations of optimization-based approaches by introducing global semantic information such as specular and reflective priors for more robust matching. However, they usually require the ground truth as supervision and are memory-consuming due to the explicitly built 3D cost volumes~\cite{yao2018mvsnet,yan2020dense,chen2019point}. To tackle these two problems, methods combining implicit representation and differentiable rendering have attracted wide attention. These methods benefit from both the flexibility of the implicit representation and the authenticity of the differentiable rendering to enable an effective reconstruction of a single scene \cite{yariv2020multiview,mildenhall2020nerf}. On the other hand, these methods do not perform well in the human head reconstruction scenario. The poor performance is mainly due to two reasons: first, with the relatively simple textures and highly complex structures of hair regions, geometric reconstruction is much more difficult. Second, due to the relatively few image inputs and various illumination conditions across different views, the reconstructed results often contain much noise, thus more prior knowledge of head modeling is needed.

In this paper, we combine the implicit differentiable rendering method~\cite{yariv2020multiview} with several head priors to reconstruct a high-fidelity head model from a few multi-view images. To enable dense reconstruction from sparse inputs and alleviate noise, we design a neural network structure that utilizes facial prior knowledge, head semantic segmentation information, and hair orientation maps as constraints for improving the reconstruction robustness and accuracy. First, we adopt an optimization-based method to recover the coarse 3D facial geometry and initial camera parameters from input images. This information provides a reasonable initial 3D structure for the implicit space. Then, we adopt the 2D semantic segmentation information obtained using a face parsing model \cite{faceparsing} to reduce the reconstruction noise. Finally, 2D orientation maps of hair regions are extracted using oriented filters and are used to guide the orientation of 3D hair in the following reconstruction process. The 3D orientation is directly computed from the implicit neural representation with a curvature-based method. Using this auxiliary information, we estimate the head geometry as the zero-level set of the signed distance field represented using an implicit neural rendering network. At last, a high-fidelity 3D head model is recovered via the Marching Cubes algorithm \cite{lorensen1987marching}. Fig.~\ref{fig:teaser} shows the overview of our method. Comprehensive experiments demonstrate that our method can reconstruct more accurate head models from relatively few image inputs, compared with state-of-the-art multi-view stereo reconstruction methods. In summary, our paper makes the following contributions:

\begin{itemize}
    \item We propose a prior-guided implicit neural network structure for the reconstruction of a high-quality 3D head model from a few multi-view portrait images.
    
    \item We introduce three different head priors to improve the reconstruction accuracy and robustness. The facial prior knowledge provides the initial geometric information of the face region for the implicit space. The head semantic segmentation information helps to maintain the geometric structures of the head, and the 2D hair orientation maps improve the geometric accuracy of the hair region.
    
    \item We adopt a curvature-based method to directly compute the 3D orientation field from implicit neural representation, enabling the utilization of the guidance provided by the 2D hair orientation maps.
\end{itemize}
\section{Related work}

\begin{figure*}[h]
	\centering
	\includegraphics[width=1\textwidth]{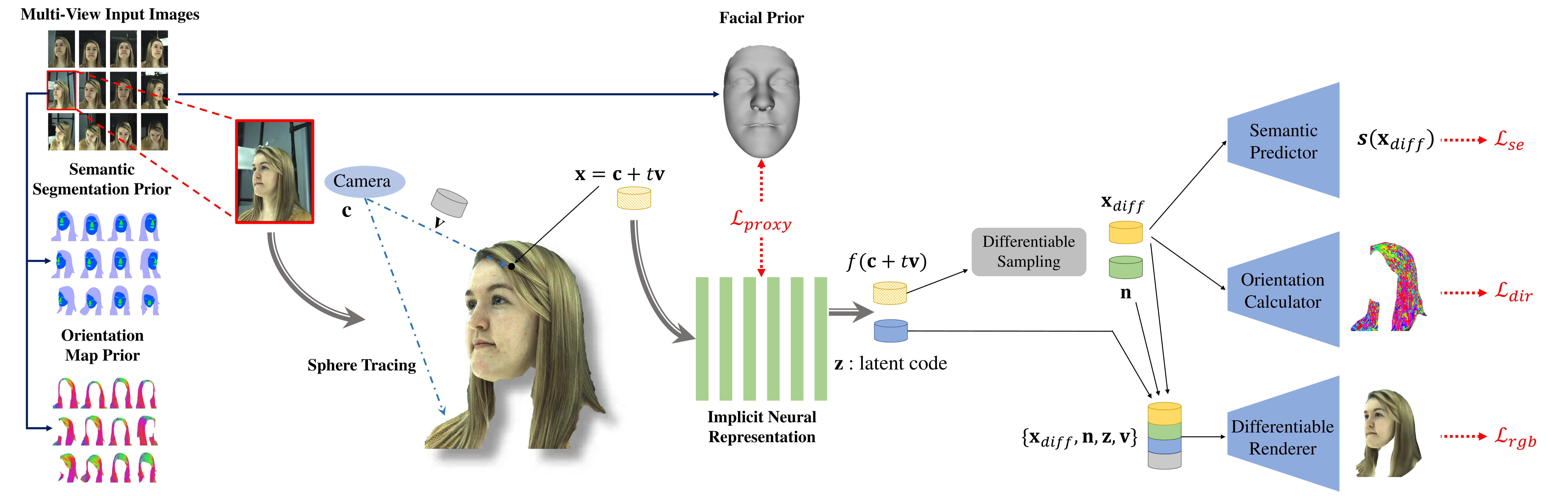}
	\caption{Overview of our approach. We first extract three different head priors from the multi-view image inputs, including the facial prior knowledge, head semantic segmentation information and hair orientation maps. Then we leverage three prior-guided loss terms in our framework to improve the reconstruction accuracy and robustness. Meanwhile, self-supervised signals come from the deviation between the input image and the rendered image. Finally, we reconstruct a high-quality 3D head model via the implicit neural rendering network.}
	\label{fig:pipeline}
\end{figure*}

In this section, we review some existing works for the following four aspects: optimization-based and deep learning-based multi-view stereo for general object reconstruction, multi-view 3D face reconstruction and 3D hair reconstruction.

\textbf{Optimization-based Multi-View Stereo.} The multi-view stereo method recovers dense 3D geometric structures of general objects from multi-view images with overlapping areas. The traditional multi-view stereo method is often combined with the Structure-from-Motion (SfM) \cite{schonberger2016structure} using basic processes of feature point extraction, camera calibration, sparse point cloud reconstruction and dense point cloud fusion. It has been expanded and improved in several reported works. Gallian \etal~\cite{galliani2015massively} extended the stereo matching method, PatchMatch Stereo \cite{bleyer2011patchmatch}, to multi-view scenarios and proposed a new diffusion-like propagation scheme. Their method can remove redundant information more effectively and is more suitable for GPU multi-core calculations. Sch{\"o}nberger \etal~\cite{schonberger2016pixelwise} adopted a method to jointly estimate the depth and normal with photometric prior information and multi-view geometric consistency in order to improve the reconstruction accuracy and efficiency. Wu \etal~\cite{wu2016fast} implemented an irregular seed-and-expand method on a GPU to solve the multi-view stereo problem and proposed a hierarchical parallel computing architecture to improve the utilization of both GPUs and CPUs. Although these optimization-based methods have achieved excellent results, they still suffer from some limitations. For example, simple object textures, highlights and a small number of input images will cause reconstruction incompleteness due to the failure of feature point extraction and matching. Thus, we use a deep learning-based method to avoid such reconstruction errors.

\textbf{Deep Learning-Based Multi-View Stereo.} With the development of convolutional neural networks, various deep learning-based methods have been proposed to solve the problem of multi-view stereo. The key insight of these methods is the construction of the 3D cost volume, that can integrate 2D image features extracted from different views, improving the accuracy of feature matching. Yao \etal~\cite{yao2018mvsnet} proposed an end-to-end deep learning network for depth inference with the help of a differentiable homography warping. Then the recurrent neural network was introduced into this problem in \cite{yao2019recurrent}. Chen \etal~\cite{chen2019point} used the coarse-to-fine strategy to predict depth residuals on 3D point clouds along the visual rays, via an edge convolution operation. A similar strategy was adopted in \cite{gu2020cascade,cheng2020deep,yang2020cost}. Although they show good performance in depth accuracy and reconstruction completeness, all of these methods are based on supervised learning and are trained on publicly available datasets obtained with calibrated cameras such as \cite{aanaes2016large,knapitsch2017tanks,yao2020blendedmvs}, resulting in poor generalization ability. To enable unsupervised learning, Khot \etal~\cite{khot2019learning} adopted the photometric consistency term in network training. Several works have improved the performance using additional strategies, such as cross-view consistency~\cite{dai2019mvs2}, normal-depth consistency, feature-wise constraints~\cite{huang2020m}, semantic segmentation information and data augmentation~\cite{xu2021self}.

However, most existing methods are memory-consuming and only estimate multi-view depth maps. Therefore, the post-processing of depth fusion such as the Poisson surface reconstruction~\cite{kazhdan2006poisson} is required to obtain a watertight 3D surface. Inspired by~\cite{yariv2020multiview}, we use an implicit neural rendering network for modeling the geometry and appearance from multi-view images. Different from~\cite{yariv2020multiview}, we introduce several head priors to improve the reconstruction accuracy and robustness of the head model.

\textbf{Multi-View 3D Face Reconstruction.} 3D face reconstruction from multi-view images has made great progress in recent years. Most existing works incorporated 3D parametric face models into this problem. Dou \etal~\cite{dou2018multi} proposed a deep recurrent neural network to fuse identity-related features to aggregate identity-specific contextual information. Wu \etal~\cite{wu2019mvf} took geometric consistency into consideration and proposed an end-to-end trainable neural network to regress a group of parameters based on the 3D Morphable Model (3DMM)~\cite{blanz1999morphable,booth20163d,paysan20093d}. Agrawal \etal~\cite{agrawal2020high} designed a model-fitting approach to deform a template model into the noisy point cloud generated from multi-view input images. Bai \etal~\cite{bai2020deep} solved the problem of non-rigid multi-view face reconstruction with appearance consistency between different views. However, most of these methods still require a large amount of 3D geometric data for supervision, and the low-dimensional parametric representation limits the accuracy of the reconstruction. Moreover, a similar idea of template model fitting is not suitable for hair region reconstruction due to a variety of hairstyles and complex geometric structures. By contrast, our method can flexibly generate various topologies with the implicit neural representation, and enables unsupervised learning with the implicit differentiable renderer.

\textbf{Multi-View 3D Hair Reconstruction.} It is difficult to recover 3D hair regions from multi-view images due to their complex geometric structures such as the self-occlusion and high similarity of different strands. Based on the fundamental work of dense 2D orientation detection \cite{paris2004capture}, various approaches have been proposed to recover high-quality hair geometries. Wei \etal~\cite{wei2005modeling} generated hair fibres in a visual hull guided by orientation information. Luo \etal~\cite{luo2012multi} used the orientation similarity to refine the initial depths. Nam \etal~\cite{nam2019strand} proposed a line-based PatchMatch stereo approach with cross-view consistency in orientation fields. The main disadvantages of these methods are reflected in two aspects. On the one hand, a large number of multi-view images are required as input. On the other hand, the capturing environment requires complicated settings such as specific lighting conditions and controllers. Some existing works on the reconstruction of complete 3D head models \cite{liang2018video,he2019data} often reconstruct the head and hair regions separately by fitting input images, and then stitch them together, usually leading to a slight gap between the two parts. In this paper, we aim to reconstruct an integrated head model from a few multi-view input images.
\section{Approach}

In this section, we present our method design that reconstructs an integrated 3D head model including the hair region from a few multi-view portrait images. When restricted to a small number of input images and the complex structure of the hair region, previous methods suffer from unreasonable shapes of facial organs, an unclear structure of head parts, and inaccurate hair reconstruction due to mismatched or missing features across different views. To tackle these problems, we introduce three head priors, including facial prior knowledge represented as a proxy face model (Sec.~\ref{subsec:proxy_reconstruction}), the head semantic segmentation information (Sec.~\ref{subsec:semantic_seg}), and the 2D hair orientation maps (Sec.~\ref{subsec:orientation}). Moreover, with the flexibility of the implicit neural representation and the authenticity of the differentiable rendering (Sec.~\ref{subsec:implicit_neural_network}), we propose a prior-guided implicit neural rendering network with elaborately designed loss terms to improve the reconstruction accuracy and robustness (Sec.~\ref{subsec:loss}). We show the pipeline of our proposed approach in Fig.~\ref{fig:pipeline} and describe each component in detail in the following subsections.

\subsection{Preliminaries}
\label{subsec:preliminaries}

\textbf{Parametric Face Model.} The 3D Morphable Model (3DMM) \cite{blanz1999morphable} describes the facial geometry and albedo in a low-dimensional subspace. Generally, this parametric model represents geometry $\mathbf{G} \in \mathbb{R}^{3n_v}$ and albedo $\mathbf{A} \in \mathbb{R}^{3n_v}$ with principal component analysis (PCA) as:
\begin{align}
\mathbf{G} &= \overline{\mathbf{G}} + \mathbf{B}_{id}\bm{\alpha}_{id} + \mathbf{B}_{exp}\bm{\alpha}_{exp}, \label{eq:3dmm_geo} \\
\mathbf{A} &= \overline{\mathbf{A}} + \mathbf{B}_{alb}\bm{\alpha}_{alb}, \label{eq:3dmm_alb}
\end{align}
where $n_v$ is the vertex number of the face model. $\overline{\mathbf{G}} \in \mathbb{R}^{3n_v}$ and $\overline{\mathbf{A}} \in \mathbb{R}^{3n_v}$ are the mean shape and albedo, respectively. $\mathbf{B}_{id} \in \mathbb{R}^{3n_v \times 199}$, $\mathbf{B}_{exp} \in \mathbb{R}^{3n_v \times 100}$ and $\mathbf{B}_{alb} \in \mathbb{R}^{3n_v \times 145}$ are the principal axes extracted from some textured face models. $\bm{\alpha}_{id} \in \mathbb{R}^{199}$, $\bm{\alpha}_{exp} \in \mathbb{R}^{100}$ and $\bm{\alpha}_{alb} \in \mathbb{R}^{145}$ are the corresponding coefficient parameters characterizing an individual textured 3D face. In this paper, we refer to the Basel Face Model 2017 \cite{gerig2018morphable} for $\mathbf{B}_{id}$ and $\mathbf{B}_{exp}$, and the diffuse albedo model of AlbedoMM \cite{smith2020morphable} for $\mathbf{B}_{alb}$. We use the 3DMM to build a proxy face model and treat it as the facial prior knowledge for further network training.

\textbf{Camera Model.} We use the standard perspective camera model to project a point in 3D space onto the image plane, as described by:
\begin{equation}
\label{eq:camera}
\mathbf{q} = \bm{\Pi}(\mathbf{R}\mathbf{V} + \mathbf{t}),
\end{equation}
where $\mathbf{q} \in \mathbb{R}^{2}$ is the projection of point $\mathbf{V} \in \mathbb{R}^{3}$, $\bm{\Pi}: \mathbb{R}^{3} \rightarrow \mathbb{R}^{2}$ is the perspective projection, $\mathbf{t} \in \mathbb{R}^3$ is the translation vector and $\mathbf{R} \in \mathbb{R}^{3 \times 3}$ is the rotation matrix obtained from Euler angles. For convenience, we represent a group of camera parameters as $\bm{\tau} = \left\{ \mathbf{R}, \mathbf{t} \right\}$ and compute the camera centre as $\mathbf{c} = \mathbf{c}(\bm{\tau}) = -\mathbf{R}^T\mathbf{t}$.

\subsection{Implicit Neural Rendering Network}
\label{subsec:implicit_neural_network}

We implicitly model the head geometry with a multi-layer perceptron (MLP) $f$ that predicts the signed distances of points in the canonical space. Under this representation, the head geometry can be extracted as the zero-level set of the neural network $f$:
\begin{equation}
\mathcal{H}_{\bm\beta} = \left\{ \mathbf{x} \in \mathbb{R}^3 \ | \ f(\mathbf{x} \ | \ \bm{\beta}) = 0 \right\},
\end{equation}
where $\bm{\beta} \in \mathbb{R}^{m}$ is the trainable parameter of $f$. To enable the differentiable rendering of the head part, inspired by~\cite{yariv2020multiview}, we use another MLP $g$ with trainable parameters $\bm{\zeta} \in \mathbb{R}^n$ to predict the appearance of points on the head model. Specifically, given a pixel, indexed by $p$, associated with some input image, let $\mathbf{c}_{p} = \mathbf{c}_p(\bm{\tau})$ denote the centre of the respective camera, with $\mathbf{v}_p = \mathbf{v}_{p}(\bm{\tau})$ as the viewing direction (i.e., the vector pointing from $\mathbf{c}_{p}$ towards $p$). Then, $R_{p} = \{\mathbf{c}_{p} + t\mathbf{v}_{p} | t \geq 0 \}$ denotes the ray through pixel $p$. Let $\mathbf{x}_{p}$ be the first intersection of the ray $R_{p}$ and the surface $\mathcal{H}_{\beta}$ that can be efficiently computed with the sphere tracing algorithm~\cite{hart1996sphere,jiang2020sdfdiff}. Then, the rendered colour of pixel $p$ can be expressed as:
\begin{equation}
\label{eq:network_g}
\mathcal{A}_{\bm\beta, \bm\zeta, \bm\tau}\left( p \right) = g\left( \mathbf{x}_{p}, \mathbf{v}_{p}, \mathbf{n}_{p}, \mathbf{z}_p \ | \ \bm{\zeta} \right),
\end{equation}
where $\mathbf{n}_{p} \in \mathbb{R}^3$ is the normal vector of $\mathbf{x}_{p}$ derived as the gradient of $f$ at $\mathbf{x}_{p}$. $\mathbf{z}_{p} \in \mathbb{R}^{256}$ is a global geometry feature vector which is described in Sec.~\ref{subsec:implementation}. Different from \cite{yariv2020multiview}, we use facial prior knowledge, head semantic segmentation information and hair orientation maps to enable robust and accurate reconstruction from a few input images.

\subsection{Proxy Model}
\label{subsec:proxy_reconstruction}

\begin{figure}[b]
	\includegraphics[width=0.48\textwidth]{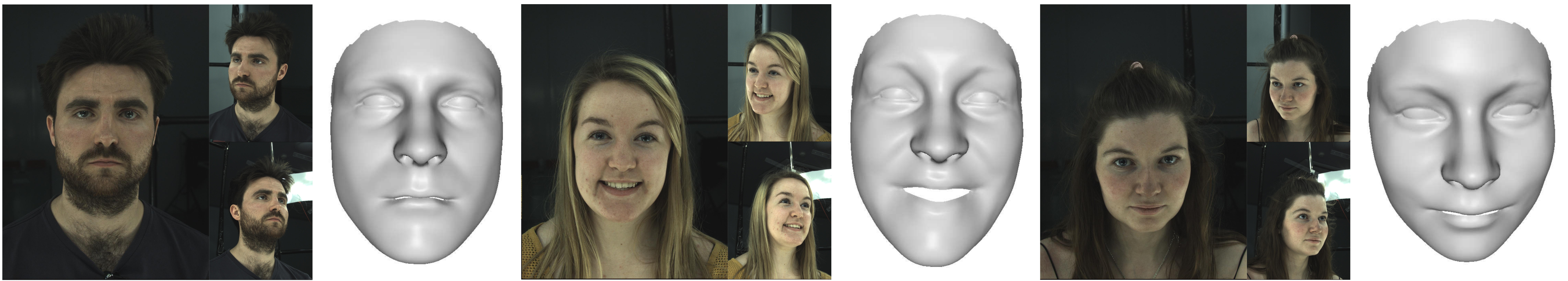}
	\caption{Some results of the proxy model reconstruction. We show three samples of all $12$ multi-view portrait images on the left and their corresponding proxy face models on the right.}
	\label{fig:proxy}
\end{figure}

To ensure a reasonable facial shape, we introduce the facial prior knowledge to guide the reconstruction in the facial region. We refer to 3DMM, the widely used PCA model of 3D faces, for the facial prior knowledge. Given $N$ multi-view images as input, we adopt an optimization-based inverse rendering method with photometric consistency to estimate the 3DMM parameters ($\bm{\alpha}_{id}$, $\bm{\alpha}_{exp}$, $\bm{\alpha}_{alb}$), lighting conditions and camera parameters $\bm\tau$. We approximate the human face as a Lambertian surface and simulate global illumination using spherical harmonics (SH) basis functions. Under this assumption, the imaging formula for vertex $i$ can be expressed as:
\begin{equation}
\label{eq:image_formula}
\mathbf{I}\left(\mathbf{n}_i, \mathbf{a}_i \ | \ \bm{\gamma} \right) = \mathbf{a}_i \cdot \left(\bm{\gamma} \cdot \bm{\phi}\left(\mathbf{n}_i\right)\right),
\end{equation}
where $\mathbf{a}_i \in \mathbb{R}^{3}$ is the albedo in the RGB colour space, $\bm{\phi}\left(\mathbf{n}_i\right) \in \mathbb{R}^{B^2}$ is the SH basis computed with the vertex normal $\mathbf{n}_i \in \mathbb{R}^3$ and $\bm{\gamma} \in \mathbb{R}^{B^2}$ is the SH coefficients. In the following we only use the first three bands ($B = 3$) of the SH basis. To estimate parameters $\mathcal{X} = \left\{ \bm{\alpha}_{id}, \bm{\alpha}_{exp}, \bm{\alpha}_{alb}, \bm{\gamma}_j, \bm\tau_j \right\}, \ \left( j = 1, 2, ..., N \right)$, we solve the following optimization problem:
\begin{equation}
\label{eq:target}
\min\limits_{\mathcal{X}} E = \min\limits_{\mathcal{X}} \ w_{ph}E_{photo} + w_{l}E_{land} + w_{r}E_{reg},
\end{equation}
where $E_{photo}$ is the photometric consistency term, $E_{land}$ is the landmark term and $E_{reg}$ is the regularization term. $w_{ph}, w_l$ and $w_r$ are their corresponding weights and are empirically set as 80.0, 5.0 and 20.0, respectively. The photometric consistency term, penalizing the deviation between the observed intensity $\mathbf{I}_{in}^j$ from the input image and the rendered intensity $\mathbf{I}^j$ with Eq.~\ref{eq:image_formula} , is described as:
\begin{equation}
E_{photo} = \sum_{j = 1}^{N} \displaystyle\frac{1}{|\mathcal{M}_{f}^j|}\left\| \mathbf{I}_{in}^j - \mathbf{I}^j \right\|_F^2,
\end{equation}
where $\mathcal{M}_{f}^j$ contains all pixels covered by the face region of view $j$. Since facial landmarks contain the structural information of the human face, we design the landmark term as shown below to measure how close the projections of 3D landmarks are to the corresponding landmarks on images:
\begin{equation}
E_{land} = \displaystyle\frac{1}{|\mathcal{L}|} \sum_{j = 1}^{N} \sum_{i \in \mathcal{L}} \left\| \mathbf{q}_{i}^j - \bm{\Pi}\left( \mathbf{R}_j\mathbf{V}_i + \mathbf{t}_j \right) \right\|_2^2,
\end{equation}
where $\mathcal{L}$ is the landmark index set, $\mathbf{V}_i$ is a 3D landmark and $\mathbf{q}_i^j$ is its corresponding landmark position on image $j$ detected using \cite{bulat2017far}. $\mathbf{R}_j$ and $\mathbf{t}_j$ are camera parameters in Eq.~\ref{eq:camera} . Finally, to ensure that the parameters of the fitted parametric face model are plausible, we introduce the regularization term as:
\begin{equation}
E_{reg} = \sum_{i = 1}^{199} \left( \displaystyle\frac{\alpha_{id, i}}{\sigma_{id, i}} \right)^2 + \sum_{i = 1}^{100} \left( \displaystyle\frac{\alpha_{exp, i}}{\sigma_{exp, i}} \right)^2 + \sum_{i = 1}^{145} \left( \displaystyle\frac{\alpha_{alb, i}}{\sigma_{alb, i}} \right)^2,
\end{equation}
where $\bm{\sigma}_{id}$, $\bm{\sigma}_{exp}$ and $\bm{\sigma}_{alb}$ are standard deviations in the corresponding principal directions.

When solving Eq.~\ref{eq:target}, we assume that all views share the same 3DMM parameters $\bm{\alpha}_{id}, \bm{\alpha}_{exp}, \bm{\alpha}_{alb}$ while the lighting parameters $\bm{\gamma}$ and camera parameters $\bm\tau$ are view-dependent. We first initialize the 3DMM parameters and lighting parameters to zero, then set the rotation matrices as unit matrices, and let the translation vectors be zero vectors. Then, we solve this optimization problem using the gradient descent method until convergence. In this paper, we mainly take $N$ as 3, 12 and 30, with their respective optimization times of approximately 36.48s, 65.90s and 119.17s on a single GeForce RTX 2080 Ti GPU. Fig.~\ref{fig:proxy} contains some reconstruction results of 12 multi-view input images.

\subsection{Semantic Segmentation}
\label{subsec:semantic_seg}

To avoid unreasonable facial organ shapes and unclear head structures, we introduce head semantic segmentation information to guide head model reconstruction. We use the face parsing network in MaskGAN \cite{faceparsing} to extract the head semantic segmentation information for all of the multi-view portrait images. MaskGAN treats the semantic segmentation task as an image-to-image translation problem, and achieves the goal of face manipulation via semantic-based mapping. It adopts a similar network architecture to Pix2PixHD \cite{wang2018high} and is trained on the CelebA-HD \cite{karras2017progressive} and CelebAMask-HQ \cite{faceparsing} datasets at the resolution of $512 \times 512$. With the pretrained model, we extract the head semantic segmentation information for all multi-view portrait images, including the hair, face, eyes, eyebrows, nose and lips. The extraction time of a single input image is approximately 0.21s on a single GeForce RTX 2080 Ti GPU. Using this head semantic segmentation information, we design a semantic neural network $s$ with trainable parameters $\bm\delta \in \mathbb{R}^h$ to predict the probabilities of each part of the segmentation at a given point $\mathbf{x} \in \mathbb{R}^3$, and a loss term to guide the probability prediction as described in Sec.~\ref{subsec:loss}.

\begin{figure}[t]
	\includegraphics[width=0.48\textwidth]{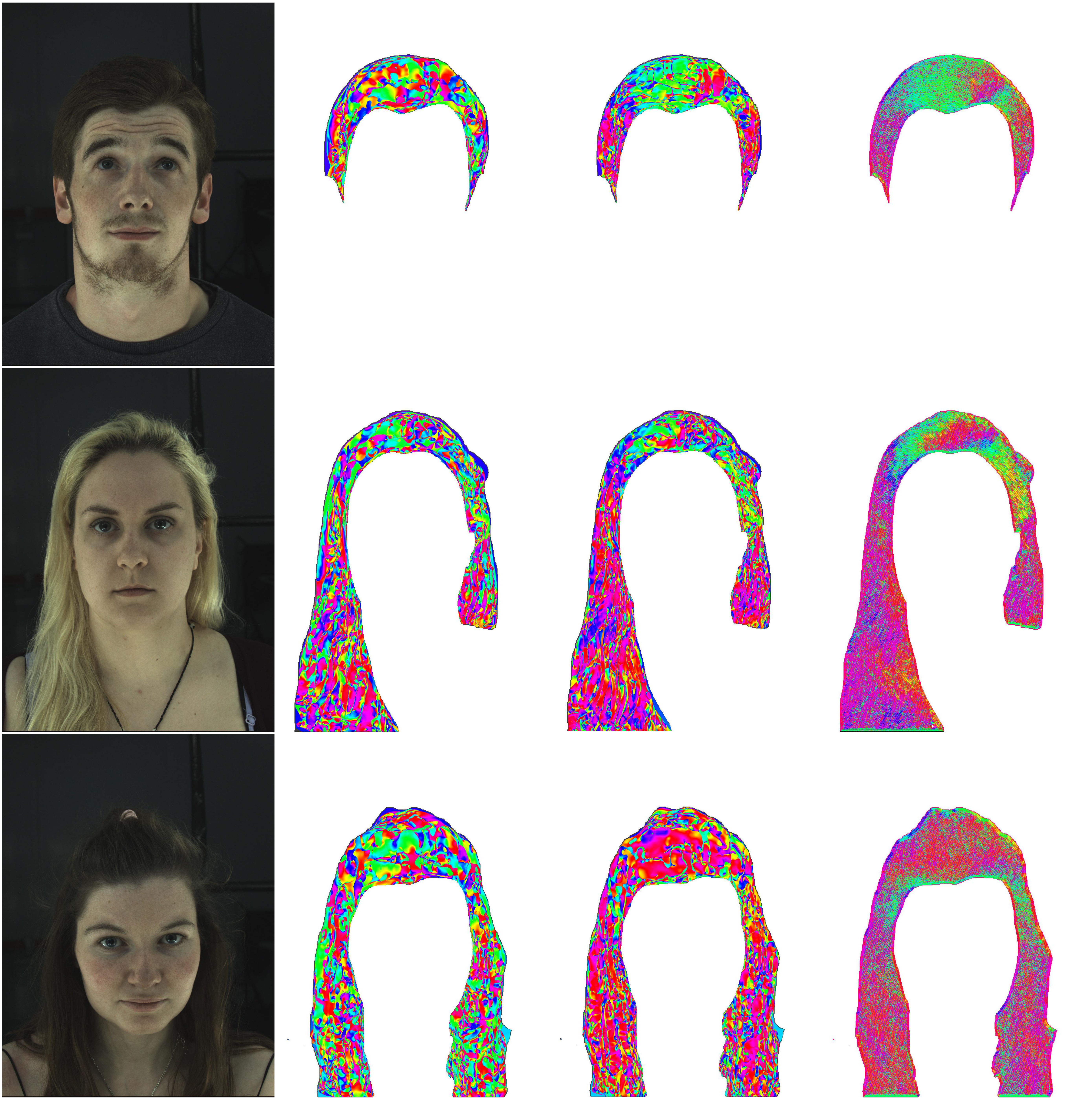}
	\caption{Some results of hair orientation fields. From left to right: the front view input images, initial 3D hair orientation fields, optimized 3D hair orientation fields using Eq.~\ref{eq:lossdir} and detected 2D orientation maps.}
	\label{fig:orientation}
\end{figure}

\subsection{Hair Orientation}
\label{subsec:orientation}

To handle the relatively simple textures and highly complex structures, we guide the hair region reconstruction with the 2D hair orientation maps to improve its reconstruction accuracy.

\textbf{2D Orientation Detection.} For a given hair image $\mathbf{I}_h$ segmented from the corresponding portrait image $\mathbf{I}_{in}$ with a hair mask $\mathcal{M}_h$, we can detect its pixel-wise orientation information with a series of oriented filters $\left\{ \mathbf{\mathcal{K}_{\theta}} \right\}$, where the filter kernel $\mathcal{K}_{\theta}$ is designed to detect the orientation at angle $\theta$. Assuming that the response of kernel $\mathcal{K}_{\theta}$ at a given pixel, indexed by $p$, is $F_{\theta}(p) = \left| \mathcal{K}_{\theta} * \mathbf{I}_h \right|\left(p \right)$, then the detected orientation at $p$ is given by
\begin{equation}
	\label{eq:2Ddir}
	\theta^* = \argmax\limits_{\theta} F_{\theta}(p).
\end{equation}
Following the fundamental work of \cite{paris2004capture} we use a filter set of 32 even-symmetric Gabor kernels \cite{jain1991unsupervised} to detect the orientation information, constructed using a Gaussian envelope to modulate an oriented sinusoidal plane wave, with $\theta \in \left[ 0, \pi \right]$:
\begin{equation}
	\mathcal{K}_{\theta}\left( u, v \right) = \exp\left( -\displaystyle\frac{1}{2} \left( \displaystyle\frac{\hat{u}^2}{\sigma_u^2} + \displaystyle\frac{\hat{v}^2}{\sigma_v^2} \right) \right)\cos\left( \displaystyle\frac{2\pi\hat{u}}{\lambda} \right),
\end{equation}
where
\begin{equation*} 
	\hat{u} = u\cos\theta + v\sin\theta, \ \hat{v} = -u\sin\theta + v\cos\theta, \\
\end{equation*}
and $(u,v)$ are the coordinates relative to $p$ of a pixel within a filtering window centred at $p$. $\sigma_u$, $\sigma_v$ are adjustable parameters that control the standard deviations of the Gaussian envelope, and $\lambda$ is a tuned parameter to adjust the period of the sinusoidal plane wave. Finally, the unit detected 2D orientation vector can be computed as $\mathbf{d}_p = (-\sin\theta^*, \cos\theta^*)$. It takes approximately 5.69s to detect the 2D hair orientations of a single input image on an I9-7900X CPU. Some results are shown in Fig.~\ref{fig:orientation}.

\textbf{3D Orientation Calculation.} According to the study of differentiable geometry, the bending degree of a 3D surface at a given point can be described by the principal curvatures and the bending direction can be described by the corresponding principal directions. Given a point $\mathbf{x} \in \mathcal{H}_{\beta}$, its normal vector can be computed by:
\begin{equation}
	\mathbf{n} = \bigtriangledown_{\mathbf{x}}f\left( \mathbf{x} \ | \ \bm{\beta} \right).
\end{equation}
Let $\mathbf{H} \in \mathbb{R}^{3 \times 3}$ denote the Jacobian matrix of $\mathbf{n}$ relative to $\mathbf{x}$, or equivalently the Hessian matrix of $f$. Then by differentiating the relation $\left< \mathbf{n}, \mathbf{n} \right> = 1$, we obtain $\mathbf{H} \cdot \mathbf{n} = \mathbf{0}$, which gives us one eigenvalue $0$ and its corresponding eigenvector $\mathbf{n}$ of $\mathbf{H}$. According to Rodrigues' formula the other two eigenvalues of $\mathbf{H}$ are the principal curvatures of $\mathcal{H}_{\bm\beta}$ at point $\mathbf{x}$, with the two corresponding eigenvectors as the principal directions \cite{mayost2014applications}. Thus we can describe the 3D orientation $\mathbf{D}_{\mathbf{x}}$ at the point $\mathbf{x}$ as the corresponding unit eigenvector of the second smallest absolute eigenvalue of the Hessian matrix $\mathbf{H}$.

\subsection{Loss}
\label{subsec:loss}

Guided by this head prior knowledge, we first leverage the three prior-related loss terms in our framework to overcome problems caused by a small number of multi-view input images and the highly complex geometric structures of hair regions. Meanwhile, to utilize the self-supervised signals provided by the multi-view images and guarantee the SDF network $f$ to be a signed distance function, we also use two consistency loss terms and one regularization loss term adopted in~\cite{yariv2020multiview} during the network training. In the following, we first describe the symbols used in the loss computation and then introduce all of the loss terms in detail.

Our networks consist of $4$ trainable parameters, including $\bm\beta$ used in the SDF network $f$, $\bm\zeta$ used in the rendering network $g$, $\bm\delta$ used in the semantic prediction network $s$, and the camera parameters $\bm\tau$. Let $p$ be a sampling pixel from a given input image. The viewing direction $\mathbf{v}_{p} = \mathbf{v}_{p}(\bm\tau)$ is the vector pointing from the camera centre $\mathbf{c}_p = \mathbf{c}_{p}(\bm\tau)$ towards $p$ and $R_{p} = \{\mathbf{c}_{p} + t\mathbf{v}_{p} | t \geq 0 \}$ is the ray through pixel $p$. Then the first intersection of $R_{p}$ and the implicit head surface $\mathcal{H}_{\bm\beta}$ is denoted by $\mathbf{x}_{p}$ which is obtained by the sphere tracing algorithm~\cite{hart1996sphere, jiang2020sdfdiff}. With these variables, the loss terms are computed as follows.

\para{Facial Proxy Term.} With facial prior knowledge obtained in Sec.~\ref{subsec:proxy_reconstruction}, we design a facial prior term to provide the initial geometric information of the face region for the implicit neural head model as:
\begin{equation}
	\label{eq:lossprior}
	\mathcal{L}_{proxy}(\bm\beta) = \displaystyle\frac{1}{\left| \mathcal{V} \right|}\sum_{\mathbf{V} \in \mathcal{V}}\left| f\left( \mathbf{V} \ | \ \bm{\beta} \right) \right|,
\end{equation}
where $\mathcal{V}$ is the collection of sampling points on the proxy face model. 

\para{Head Semantic Term.} The semantic term is used to reduce the impact of the noise on the head geometric structures caused by the small number of multi-view input images:
\begin{equation}
	\label{eq:lossse}
	\mathcal{L}_{se}(\bm\beta, \bm\delta, \bm\tau) = \displaystyle\frac{1}{\left| \mathcal{M} \right|}\sum_{p \in \mathcal{M}}CE(s(\mathbf{x}_{p} \ | \ \bm\delta), L_{p}),
\end{equation}
where $CE$ is the cross-entropy loss. The input mask $\mathcal{M}$ contains all pixels in the 2D head region, obtained from the semantic segmentation result in Sec.~\ref{subsec:semantic_seg}. $p \in \mathcal{M}$ indicates a pixel $p$ in the input mask $\mathcal{M}$, $L_p$ is the extracted 2D semantic segmentation information in Sec.~\ref{subsec:semantic_seg} and $s(\mathbf{x}_{p} \ | \ \bm\delta)$ is the predicted semantic segmentation probability. 

\para{Hair Orientation Term.} We use the hair orientation term to estimate how close the unit detected orientation $\mathbf{d}_p \in \mathbb{R}^2$ is to the projected orientation $\mathbf{d}_{\mathbf{x}_{p}} \in \mathbb{R}^2$ of a 3D orientation $\mathbf{D}_{\mathbf{x}_{p}} \in \mathbb{R}^3$ obtained in Sec.~\ref{subsec:orientation}:
\begin{equation}
	\label{eq:lossdir}
	\mathcal{L}_{dir}(\bm\beta, \bm\tau) = \displaystyle\frac{1}{\left| \mathcal{M}_h \right|}\sum_{p \in \mathcal{M}_h}(1 - \left| \mathbf{d}^T_p\mathbf{d}_{\mathbf{x}_{p}} \right| ),
\end{equation}
where $\mathcal{M}_h$ is the input hair mask.

\para{Photometric Consistency Term.} To utilize the self-supervised signals provided by the multi-view images, we implement the photometric consistency term, that penalizes the deviation between the input image $\mathbf{I}_{in}$ and the rendered image $\mathcal{A}_{\bm\beta, \bm\zeta, \bm\tau}$ from Eq.~\ref{eq:network_g}, as:
\begin{equation}
	\mathcal{L}_{rgb}(\bm\beta, \bm\zeta, \bm\tau) = \displaystyle\frac{1}{\left| \mathcal{M} \right|}\sum_{p \in \mathcal{M}} \left| \mathbf{I}_{in}(p) - \mathcal{A}_{\bm\beta,\bm\zeta, \bm\tau}\left( p \right) \right|.
\end{equation}

\para{Contour Consistency Term.} Similar to~\cite{yariv2020multiview}, we use a mask term to ensure the contour-consistency of the reconstructed 3D head model and the 2D input mask:
\begin{equation}
	\mathcal{L}_{mask}(\bm\beta, \bm\tau) = \displaystyle\frac{1}{\alpha\left| \mathcal{P} \right|}\sum_{p \in \mathcal{P}} CE\left( O_p, S_{\alpha,\bm\beta}(p) \right),
\end{equation}
where $\mathcal{P}$ is the set of all sampling pixels, $\alpha > 0$ is a parameter, $O_p$ is an indicator function identifying whether pixel $p$ is in the input mask $\mathcal{M}$, and $S_{\alpha,\bm\beta}$ is a differentiable approximation of whether $p$ is occupied by the rendered object:
\begin{equation}
	S_{\alpha,\bm\beta}(p) = \textrm{sigmoid}\left( -\alpha \min_{t \geq 0} f(\mathbf{c}_{p} + t\mathbf{v}_p \ | \ \bm\beta) \right).
\end{equation}

\para{Eikonal Regularization Term.} The Eikonal regularization term enforces network $f$ to be approximately a signed distance function:
\begin{equation}
	\mathcal{L}_{e}(\bm\beta) =\mathbb{E}_\mathbf{x}\left( \left\| f(\mathbf{x} \ | \ \bm{\beta}) \right\| - 1 \right)^2.
\end{equation}

\noindent Combining all of these loss terms, our loss function has the form:
\begin{equation}
	\label{eq:lossfunc}
	\begin{aligned}
		\mathcal{L}(\bm\beta, \bm\delta, \bm\zeta, \bm\tau) & = \ w_{rgb}\mathcal{L}_{rgb}(\bm\beta, \bm\zeta, \bm\tau) + w_{m}\mathcal{L}_{mask}(\bm\beta, \bm\tau) \\
		& + w_e\mathcal{L}_e(\bm\beta) + w_{p}\mathcal{L}_{proxy}(\bm\beta) \\ 
		& + w_{s}\mathcal{L}_{se}(\bm\beta, \bm\delta, \bm\tau) + w_{d}\mathcal{L}_{dir}(\bm\beta, \bm\tau).
	\end{aligned}
\end{equation}
With this well-designed loss function, our method can recover a high-quality 3D head model extracted from the signed distance fields using the Marching Cubes algorithm \cite{lorensen1987marching}.
\section{Experiments}

\begin{figure*}[htbp]
	\includegraphics[width=0.99\textwidth]{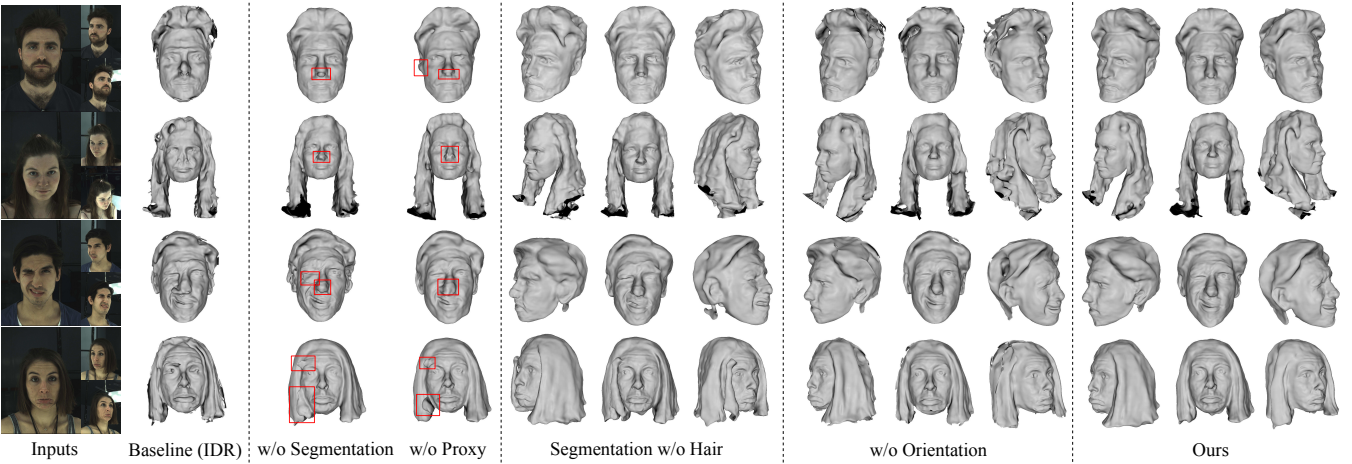}
	\caption{Ablation studies that compare the proposed method with five approaches excluding all of the head priors (baseline), the head semantic segmentation information, the facial prior knowledge, the hair segmentation information, and the hair orientation guidance. On the left, we show three input images under the front, left, and right views of all 12 multi-view input images. On the right, we show the reconstructed 3D head models of different settings.}
	\label{fig:ablation}
\end{figure*}

\begin{table*}[!t]
	\centering  
	\caption{Average geometric errors (mm) on the phone-captured data for ablation studies. We take 12 multi-view images as input and compute the average point-to-point distance from our reconstructed model to the ground truth model, as obtained with Agisoft Metashape~\cite{photoscan} from 200 multi-view images.}
	\label{tab:ablation_study}
	\centering
	\begin{tabular}{ccccccc}  
		\toprule   
		\# Input & Baseline(IDR) & w/o Segmentation & Segmentation w/o Hair & w/o Proxy & w/o Orientation & Ours \\  
		\midrule   
		12 & 2.669 & 3.048 & 2.388 & 2.345 & 2.473 & \textbf{2.301}  \\  
		\bottomrule  
	\end{tabular}
\end{table*}

\begin{figure*}[htbp]
	\includegraphics[width=0.95\textwidth]{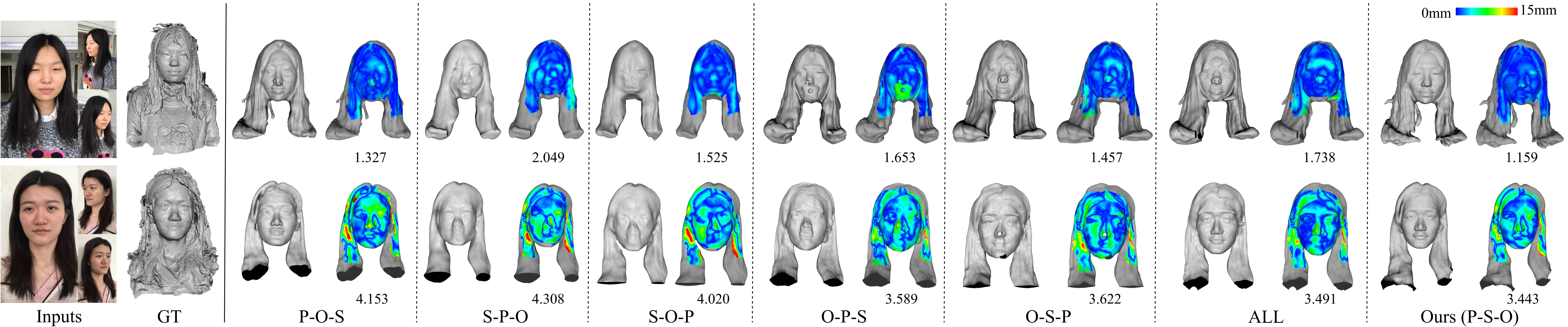}
	\caption{Quantitative comparison of different network optimization strategies on the phone-captured data. Three portrait images of all 12 multi-view input images are shown in the first column. Ground truth models, obtained with Agisoft Metashape \cite{photoscan} from 200 multi-view images, are shown in the second column. \emph{P}, \emph{S} and \emph{O} refer to the facial proxy loss term  Eq.~\ref{eq:lossprior}, the head semantic loss term Eq.~\ref{eq:lossse} and the hair orientation loss term Eq.~\ref{eq:lossdir} respectively.}
	\label{fig:loss_order}
\end{figure*}

\begin{table*}[!t]
	\centering  
	\caption{Average geometric errors (mm) on the phone-captured data for the evaluation of the loss weights. We take 12 multi-view images as input and compute the average point-to-point distance from our reconstructed model to the ground truth model, as obtained with Agisoft Metashape~\cite{photoscan} from 200 multi-view images. The symbol ``$\times$'' indicates the multiple of the loss weight change. The symbols ``+'' and ``-'' represent the increase and decrease of the geometric error change, respectively, compared to the average geometric error of 2.301mm of our loss setting in Sec.~\ref{subsec:implementation}.}
	\label{tab:loss_weights}
	\centering
	\begin{tabular}{c c|c c|c c|c c|c c|c c|c}  
		\toprule   
		\# Input & \multicolumn{2}{c}{$w_{rgb}$} & \multicolumn{2}{c}{$w_{m}$} & \multicolumn{2}{c}{$w_{e}$} & \multicolumn{2}{c}{$w_{p}$} & \multicolumn{2}{c}{$w_{s}$} & \multicolumn{2}{c}{$w_{d}$} \\  
		\midrule   
		\multirow{2}{*}{12} & $\times$0.5 & +0.072 & $\times$0.2 & +0.122 & $\times$0.1 & +0.509 & $\times$0.2 & +0.205 & $\times$0.1 & +0.145 & $\times$0.1 & +0.075  \\  
		 & $\times$20.0 & +0.405 & $\times$2.0 & +0.100 & $\times$10.0 & +0.282 & $\times$5.0 & +0.184 & $\times$10.0 & +0.184 & $\times$10.0 & +0.435  \\  
		\bottomrule  
	\end{tabular}
\end{table*}

\begin{figure}[h]
	\includegraphics[width=0.48\textwidth]{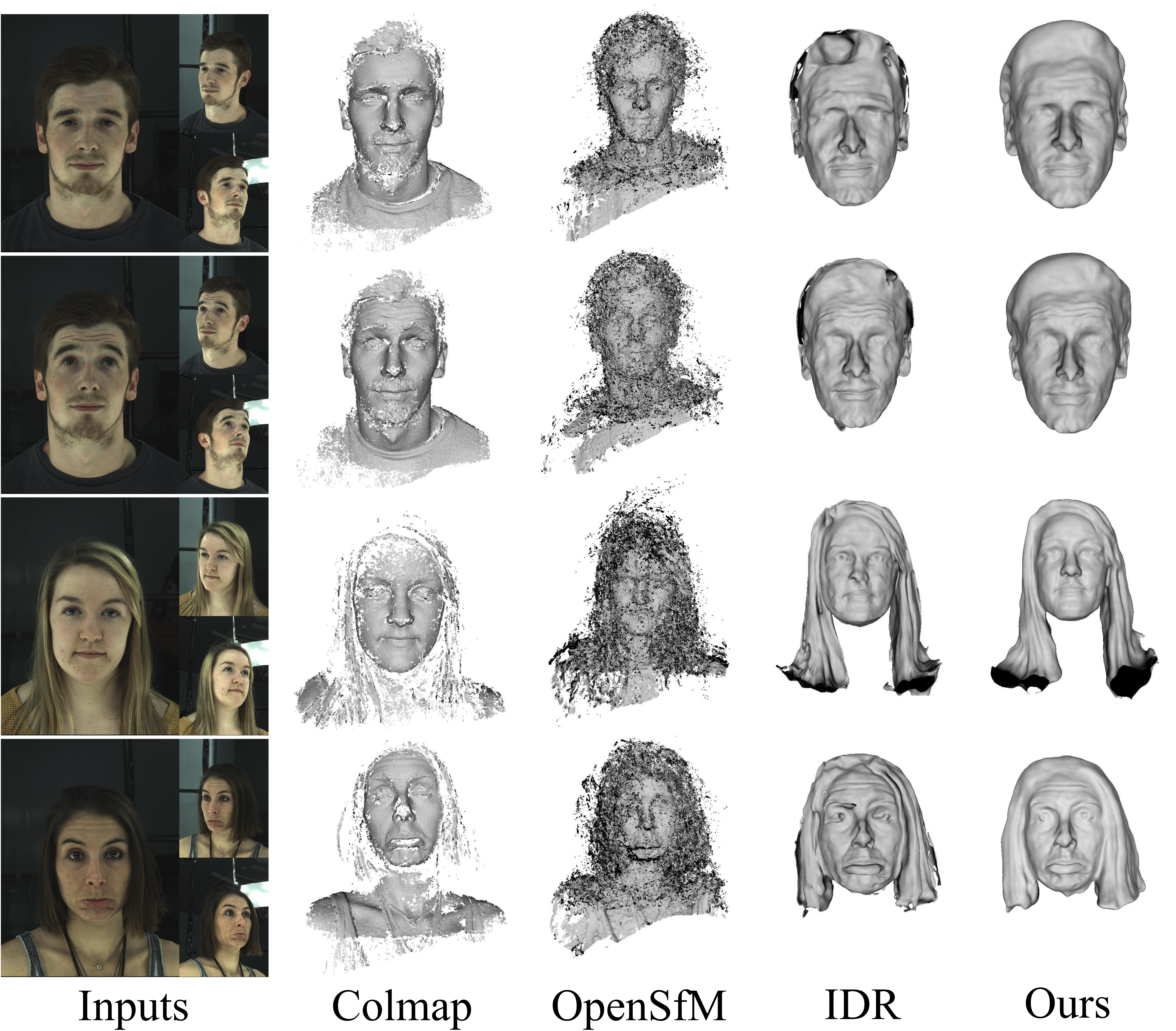}
	\caption{Qualitative comparison between Colmap \cite{schonberger2016pixelwise}, OpenSfM \cite{opensfm}, IDR \cite{yariv2020multiview} and our method. Three portrait images of all the 12 multi-view input images are shown in the first column. Compared with these methods, our approach can reconstruct more accurate and complete 3D head models.}
	\label{fig:dataset_mvs}
\end{figure}

\begin{figure*}[h]
	\includegraphics[width=0.99\textwidth]{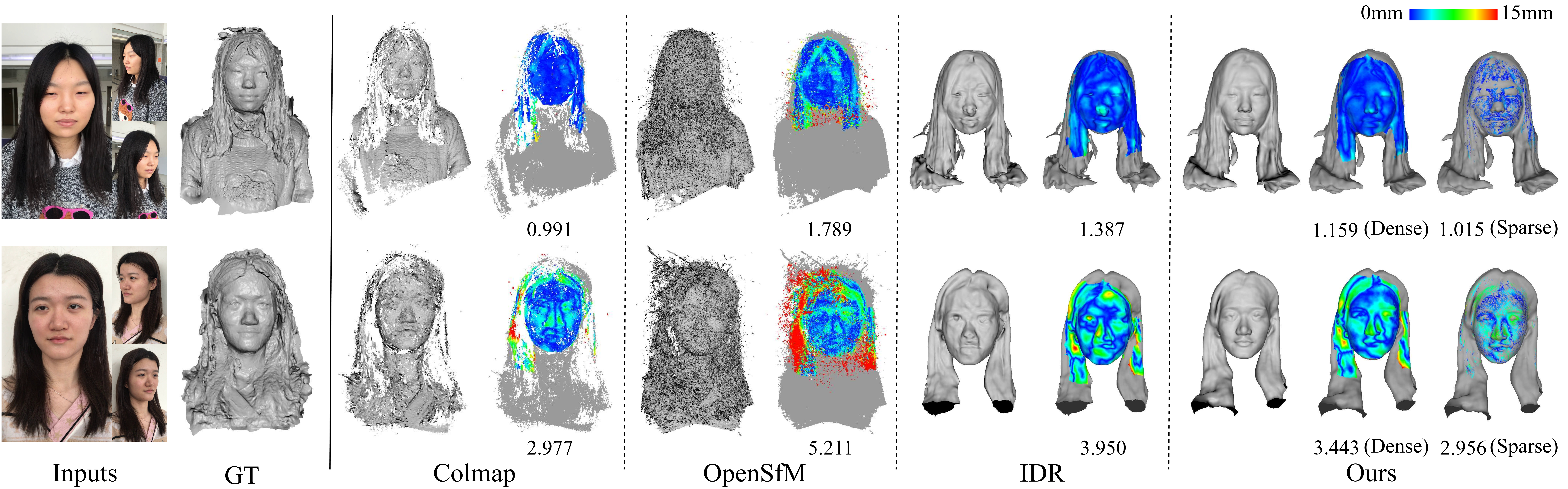}
	\caption{Quantitative comparison between Colmap \cite{schonberger2016pixelwise}, OpenSfM \cite{opensfm}, IDR \cite{yariv2020multiview} and our method on phone-captured data. Three portrait images of all 12 multi-view input images are shown in the first column. Ground truth models, obtained with Agisoft Metashape \cite{photoscan} from 200 multi-view images, are shown in the second column. The \emph{dense} error is the average point-to-point distance from our reconstructed model to the ground truth model. The \emph{sparse} error is obtained by first finding the corresponding points on our reconstructed model to Colmap's \cite{schonberger2016pixelwise} and then computing the point-to-point distance of these corresponding points to the ground truth models.}
	\label{fig:phone_mvs}
\end{figure*}

\begin{figure*}[htbp]
    \includegraphics[width=0.99\textwidth]{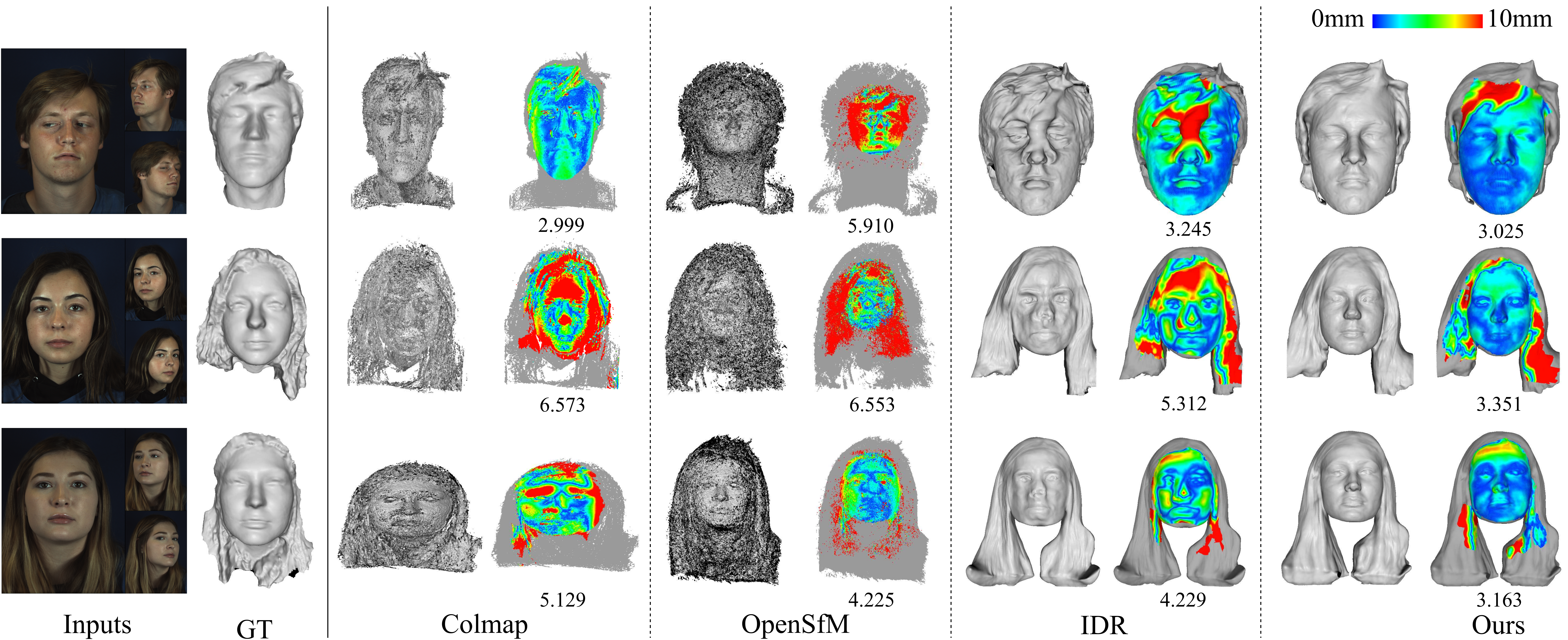}
    \caption{Reconstructed results and geometric errors of Colmap \cite{schonberger2016pixelwise}, OpenSfM \cite{opensfm}, IDR \cite{yariv2020multiview} and our method on 3 subjects of the publicly available dataset 3DFAW \cite{pillai20192nd}. Three portrait images of all 30 multi-view input images are shown in the first column, and their corresponding ground truth models, provided by 3DFAW \cite{pillai20192nd}, are shown in the second column. The geometric error is the average point-to-point distance from our reconstructed model to the ground truth model.}
    \label{fig:public_mvs}
\end{figure*}

\begin{figure}[h]
	\includegraphics[width=0.48\textwidth]{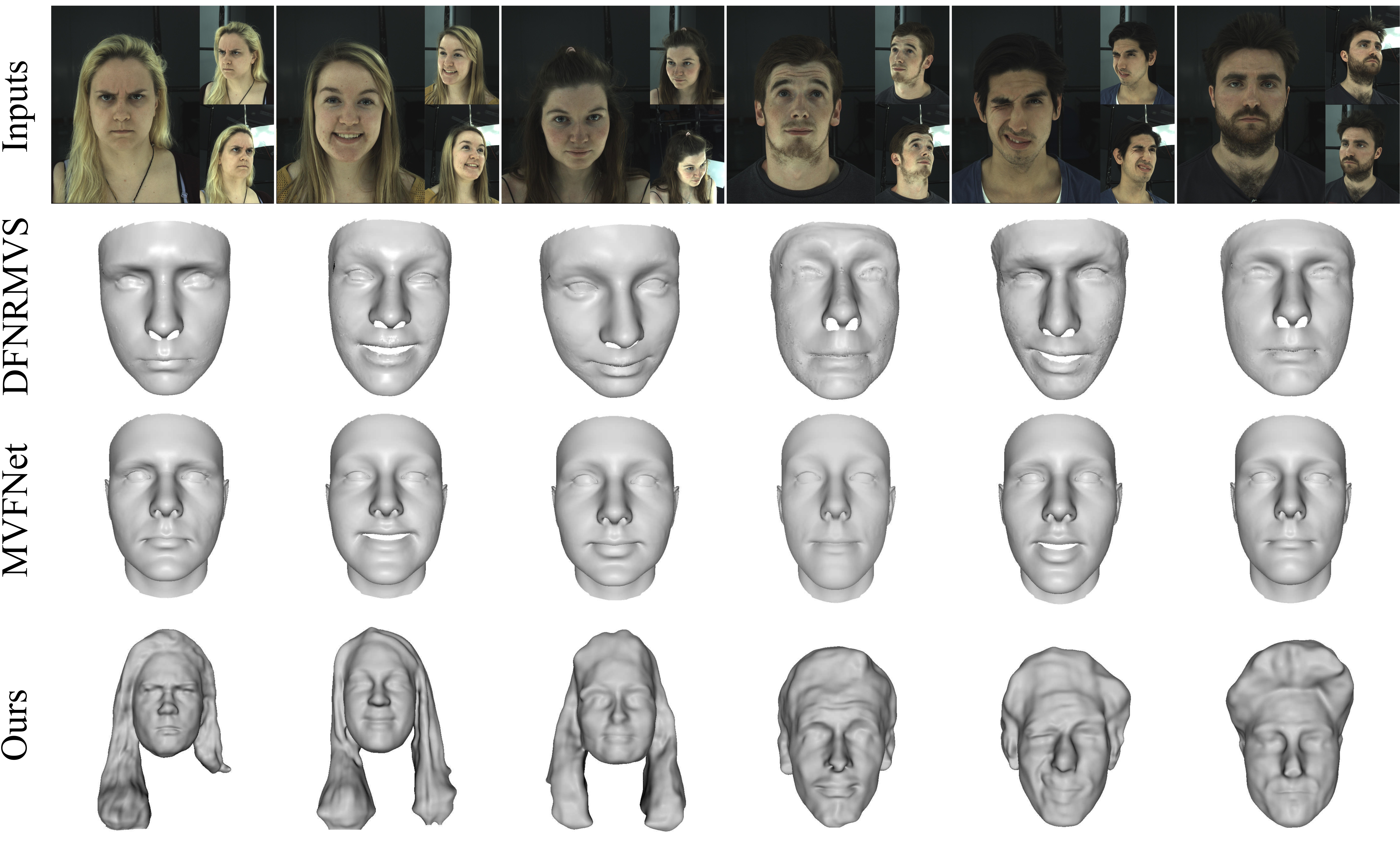}
	\caption{Qualitative comparison between DFNRMVS \cite{bai2020deep}, MVFNet \cite{wu2019mvf} and our method. All methods take three images under the front, right and left views as input. Compared with these methods our method can reconstruct more accurate face regions with fine details such as wrinkles.}
	\label{fig:dataset_mvsface}
\end{figure}

\begin{figure}[h]
	\includegraphics[width=0.48\textwidth]{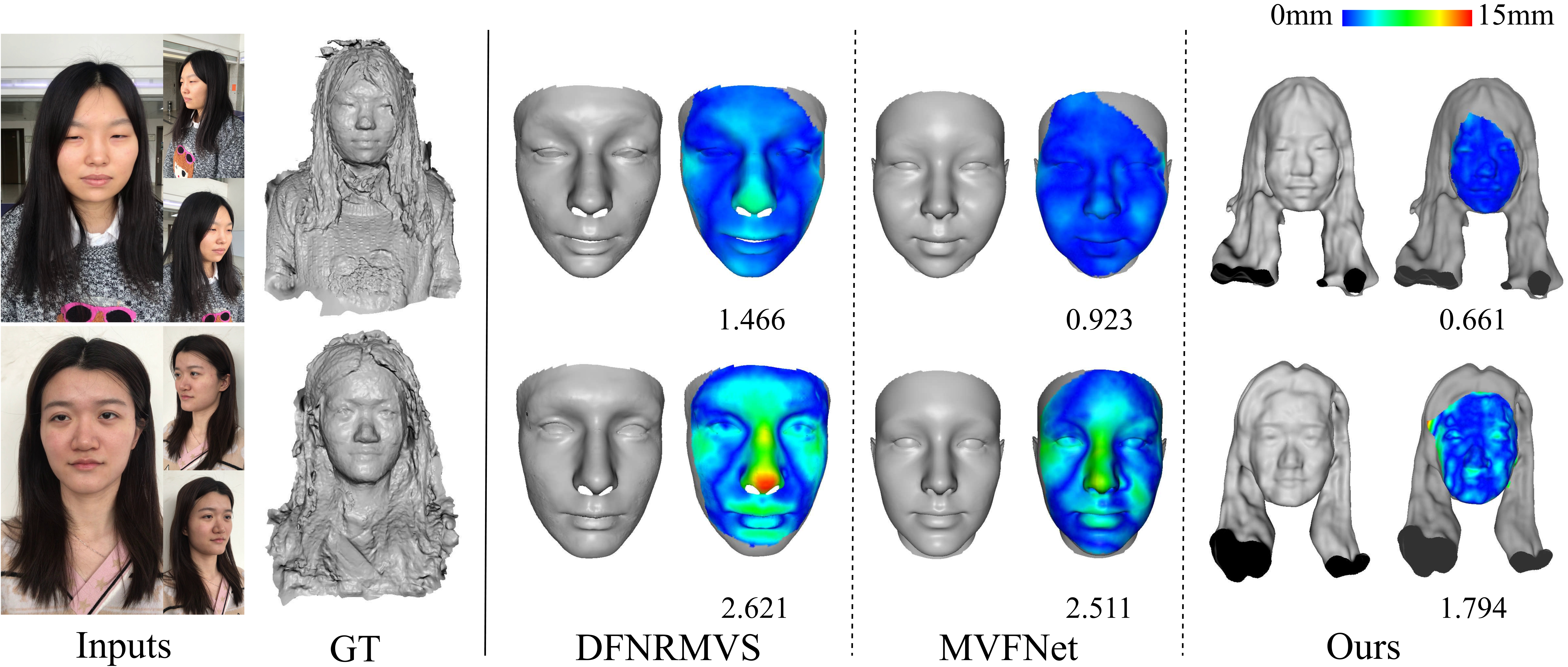}
	\caption{Reconstructed results and geometric errors of DFNRMVS \cite{bai2020deep}, MVFNet \cite{wu2019mvf} and our method on the photo-captured data. All methods take three images under the front, right and left views as input. Ground truth models, obtained with Agisoft Metashape \cite{photoscan} from 200 multi-view images, are shown in the second column.}
	\label{fig:phone_mvsface}
\end{figure}

\begin{figure}[h]
	\includegraphics[width=0.47\textwidth]{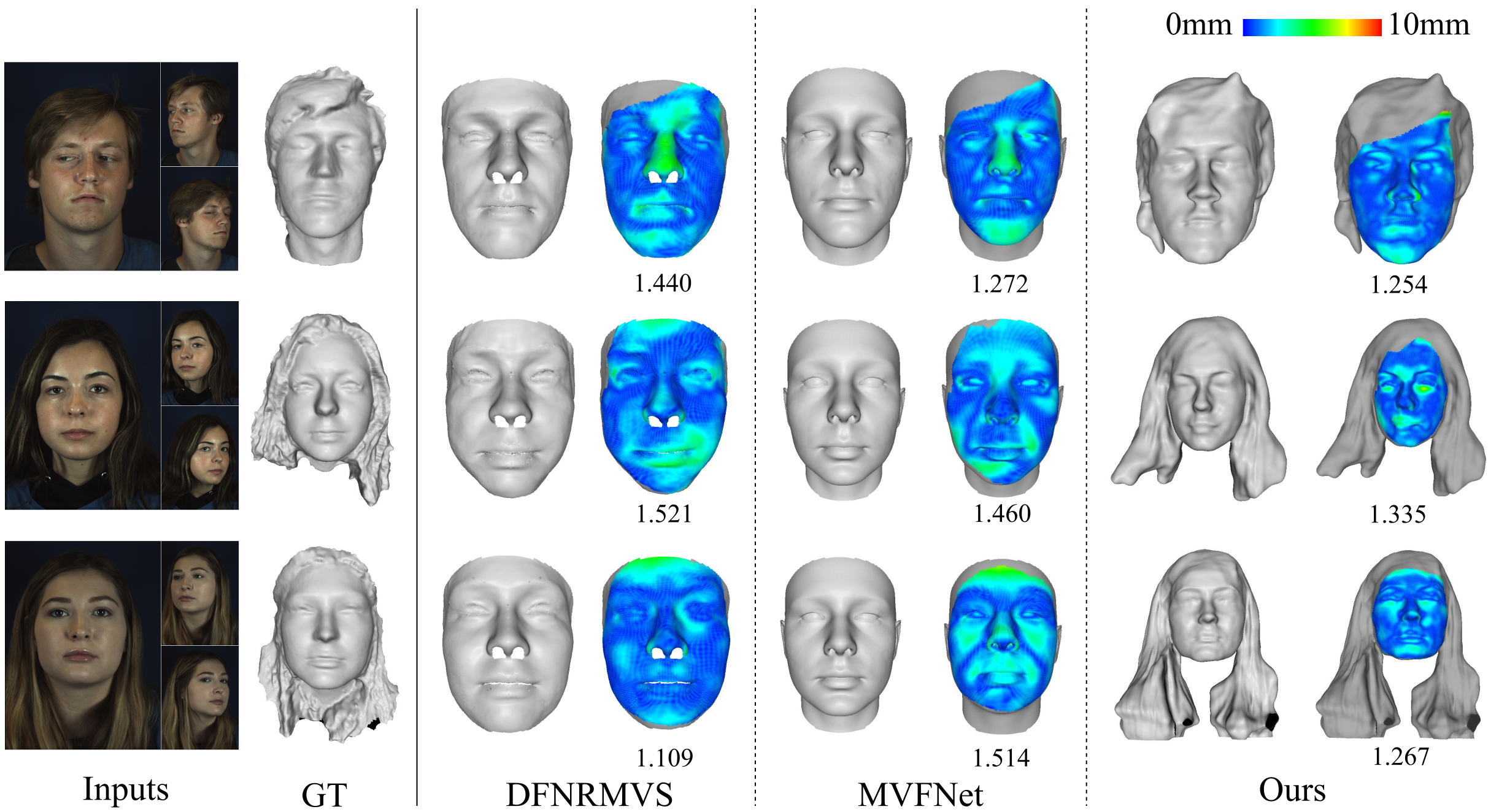}
	\caption{Quantitative comparison between DFNRMVS \cite{bai2020deep}, MVFNet \cite{wu2019mvf} and our method on 3 subjects of the publicly available dataset 3DFAW \cite{pillai20192nd}. All methods take three images under the front, right and left views as input. Ground truth models, provided by 3DFAW \cite{pillai20192nd}, are shown in the second column.}
	\label{fig:public_mvsface}
\end{figure}

\subsection{Implementation Details}
\label{subsec:implementation}

\textbf{Evaluation Data.} We evaluate our approach on three kinds of datasets with real portrait images. We first collected 12 portrait images under 12 different views using a system consisting of 12 fixed cameras with indoor lighting conditions. All of the images were captured at a resolution of $1600 \times 1200$. We also captured 2 in-the-wild portrait videos with a mobile phone ranging from 0 to 180 degrees. Then, 12 images under different views were extracted at a resolution of $1920 \times 1080$ from each video. To quantitatively measure the accuracy of our reconstruction approach, we also recovered the 3D model from 200 multi-view images extracted from each captured video as the ground truth model, using the Agisoft Metashape photogrammetry software \cite{photoscan}. These 200 multi-view images cover the range from 0 to 180 degrees with a change of 0.9 degrees between adjacent views. Finally, we evaluate our approach on the 3DFAW \cite{pillai20192nd}, a publicly available dataset containing 26 subjects, each with approximately 155 different views ranging from 0 to 180 degrees and the corresponding 3D head model. For these data, we uniformly select 30 different views at a resolution of $1392 \times 1040$ and use the corresponding 3D head models as the ground truth.

\textbf{Network Architecture.} Our approach contains three MLP networks. The implicit neural representation network $F(\mathbf{x} \ | \ \bm\beta) = (f(\mathbf{x} \ | \ \bm\beta), \mathbf{z}(\mathbf{x} \ | \ \bm\beta)), $ takes a 3D point $\mathbf{x}$ as input, and outputs a 1-dimensional signed distance value $d = f(\mathbf{x} \ | \ \bm\beta)$ and a 256-dimensional global geometry feature vector $\mathbf{z} = \mathbf{z}(\mathbf{x} \ | \ \bm\beta)$. $F$ consists of $8$ layers with hidden layers of width 512 and a skip connection from the input to the middle layer. The differentiable renderer $g(\mathbf{x},\mathbf{v},\mathbf{n},\mathbf{z} \ | \ \bm\zeta)$ consists of 4 layers with hidden layers of width 512, estimating RGB values at $\mathbf{x}$ given the viewing direction $\mathbf{v}$, the latent code $\mathbf{z}$ and the normal $\mathbf{n}$ of $\mathbf{x}$. In addition, the semantic neural network $s(\mathbf{x} \ | \ \bm\delta)$ contains 4 layers with 512-dimensional hidden layers, predicting the semantic probabilities of the given point $\mathbf{x}$.

\textbf{Optimization} We optimize our prior-guided implicit neural network for each subject. This requires the input multi-view portrait images with corresponding masks and camera parameters. In our approach, we initialize the camera parameters using the optimization-based method described in Sec.~\ref{subsec:proxy_reconstruction} and optimize them during network optimization. At each optimization epoch, we first input some randomly sampled points from the face proxy model recovered in Sec.~\ref{subsec:proxy_reconstruction} to the implicit neural representation network $F$. Meanwhile, we randomly sample some camera rays from the set of pixels in the head regions and obtain the corresponding 3D intersection points $\mathbf{x} = \mathbf{c} + t\mathbf{v}$ of each ray with the implicit surface, using the sphere tracing algorithm \cite{hart1996sphere,jiang2020sdfdiff}. Then we use the implicit neural representation network $F$ to estimate the signed distance value $d = f (\mathbf{x} \ | \ \bm{\beta})$ and the latent feature code $\mathbf{z} = \mathbf{z}(\mathbf{x} \ | \ \bm\beta)$ of each point. Due to the non-differentiability of the sphere tracing algorithm, we implement the differentiable sampling strategy used in \cite{yariv2020multiview} to ensure the differentiability of the network:
\begin{equation}
	\begin{aligned}
		\mathbf{x}_{diff} = \mathbf{x} - \displaystyle\frac{\mathbf{v}}{\bigtriangledown_{\mathbf{x}}f\left( \mathbf{x} \ | \ \bm{\beta} \right) \cdot \mathbf{v}}f\left( \mathbf{x} \ | \ \bm{\beta} \right)
	\end{aligned}
\end{equation}
where $\mathbf{c}$ is the camera centre and $\mathbf{v}$ is the viewing direction.

In our experiments, we set a batch size of 1 input image with 2048 sampling camera rays in the head region and an additional 2048 rays in the hair region. We implement our approach in PyTorch \cite{paszke2017automatic} and optimize the network parameters with the Adam solver \cite{kingma2014adam}. We optimize our networks for 3000 epochs, and sequentially introduce facial prior knowledge, head semantic segmentation information and 2D hair orientation maps into our networks every 1000 epochs. We empirically set $w_{rgb}$, $w_m$, $w_e$, $w_p$, $w_s$ and $w_d$ to 1.0, 100.0, 0.1, 1.0, 0.05 and 1.0 respectively. When introducing the 2D hair orientation maps, we set $w_{rgb}$ as 100.0 and $w_e$ as 10.0 to ensure the reconstruction accuracy.

\subsection{Ablation Study}

To validate the effect of the prior guidance, we compare the proposed method with alternative strategies that exclude some components. First, we conduct an experiment of our baseline method (IDR \cite{yariv2020multiview}) without any prior loss term. Second, we demonstrate the necessity of facial prior knowledge by conducting an experiment that excludes the facial proxy loss term Eq.~\ref{eq:lossprior} from the loss function Eq.~\ref{eq:lossfunc}. Third, we show the effectiveness of the head semantic segmentation information with another experiment that excludes the head semantic loss term Eq.~\ref{eq:lossse}. Furthermore, we study the influence of the head semantic segmentation information without the hair label by viewing the hair region as the background while keeping other labels. Finally, the guidance of 2D orientation maps is excluded from the network. Both the qualitative and quantitative comparison results for some subjects are shown in Fig.~\ref{fig:ablation} and Tab.~\ref{tab:ablation_study} respectively. It is observed that exclusion of any component will cause a drop in the performance. Specifically, facial prior knowledge and head semantic segmentation information both contribute to the reconstruction of basic geometric structures. Furthermore, the head semantic segmentation without the hair label mainly influences the facial organs. And the hair orientation guidance can improve the reconstruction accuracy of the hair regions.

\subsection{Network Optimization Evaluations}

\textbf{Evaluations on the network optimization strategy.} As described in Sec.~\ref{subsec:implementation}, we optimize our networks for 3000 epochs and sequentially introduce facial prior knowledge, head semantic segmentation information and 2D hair orientation maps into our network every 1000 epochs. To indicate the influence of the sequence of introduction of the three priors, we carry out five different experiments by exchanging the introduction sequence of the facial proxy loss term Eq.~\ref{eq:lossprior}, the head semantic loss term Eq.~\ref{eq:lossse} and the hair orientation loss term Eq.~\ref{eq:lossdir}. Meanwhile, we conduct another experiment with all three prior-related loss terms at the beginning of the network optimization. For simplicity, we refer to the facial proxy loss term  Eq.~\ref{eq:lossprior},  the head semantic loss term Eq.~\ref{eq:lossse} and the hair orientation loss term Eq.~\ref{eq:lossdir} as \emph{P}, \emph{S} and \emph{O}. Then, we quantitatively compare these results obtained using these methods for the phone-captured data. For all of the experiments, we take 12 multi-view portrait images as the input and use the same loss weights described in Sec.~\ref{subsec:implementation}. We compute the geometric errors by first applying a transformation with seven degrees of freedom (six for rigid transformation and one of scaling) to align the reconstructed head models with the ground truth models, and then computing the point-to-point distance to the ground truth models. The average geometric errors in Fig.~\ref{fig:loss_order} of \emph{P-O-S}, \emph{S-P-O}, \emph{S-O-P}, \emph{O-P-S}, \emph{O-S-P}, \emph{ALL} and our methods are 2.740mm, 3.179mm, 2.773mm, 2.621mm, 2.540mm, 2.615mm and 2.301mm, respectively. This indicates that our proposed optimization strategy can achieve the highest reconstruction accuracy.

\textbf{Evaluations on the loss weights.} As mentioned in Sec.~\ref{subsec:implementation}, we empirically set the loss weights (in Eq.~\ref{eq:lossfunc}) $w_{rgb}$, $w_m$, $w_e$, $w_p$, $w_s$ and $w_d$ to 1.0, 100.0, 0.1, 1.0, 0.05 and 1.0, respectively. To evaluate the influence of the loss weights on the reconstruction results, we carry out 12 different experiments on the photo-captured data, by reducing or increasing one loss weight at a time. All of these experiments take 12 multi-view portrait images as the input and use the same network optimization strategy described in Sec.~\ref{subsec:implementation}. Then, we compute the point-to-point distance between the reconstructed results and the ground truth models. The average geometric errors are listed in Tab.~\ref{tab:loss_weights}, indicating that our loss weight setting is appropriate.

\begin{figure*}[htbp]
	\includegraphics[width=0.99\textwidth]{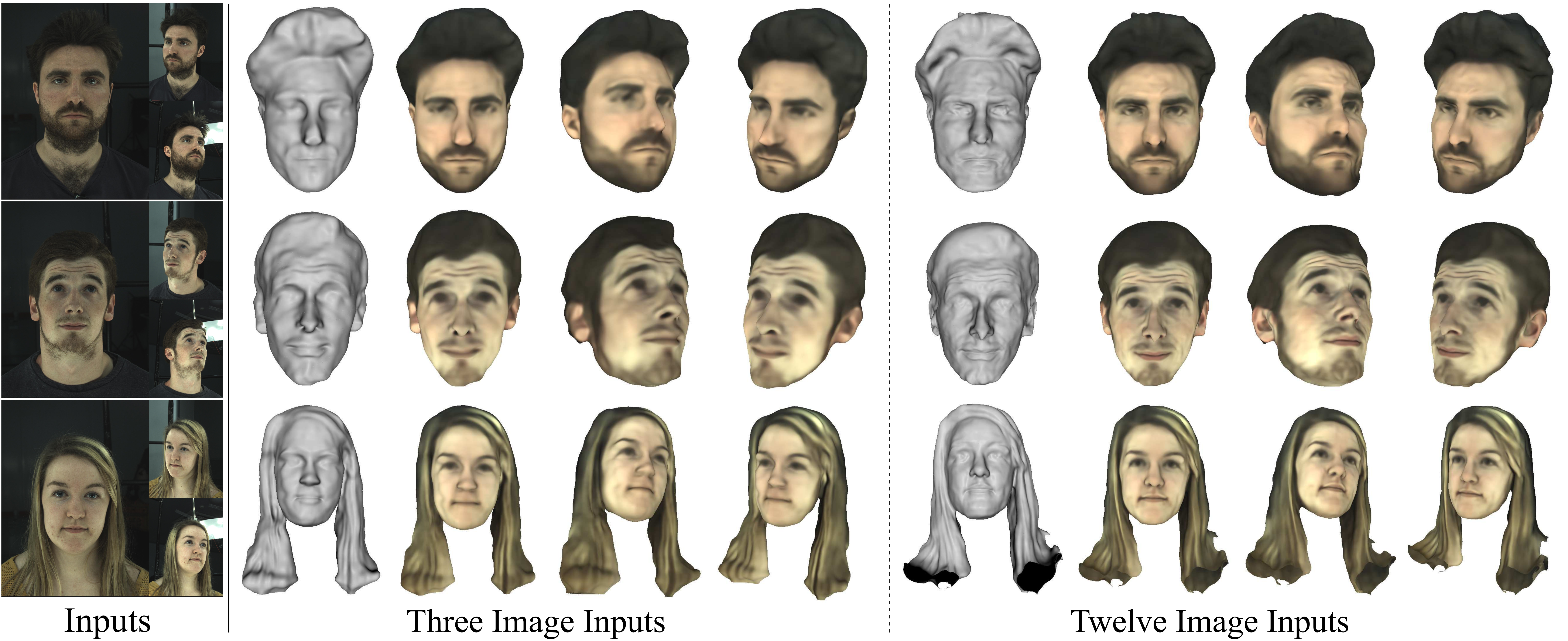}
	\caption{Qualitative results from three and twelve input images. We take three portrait images under the front, right and left views as input for the reconstruction from three input images. Another nine portrait images are used together for the reconstruction from twelve input images. Reconstructed 3D head models and their textured meshes are shown, respectively.}
	\label{fig:dataset_multi}
\end{figure*}

\begin{figure*}[htbp]
	\includegraphics[width=0.99\textwidth]{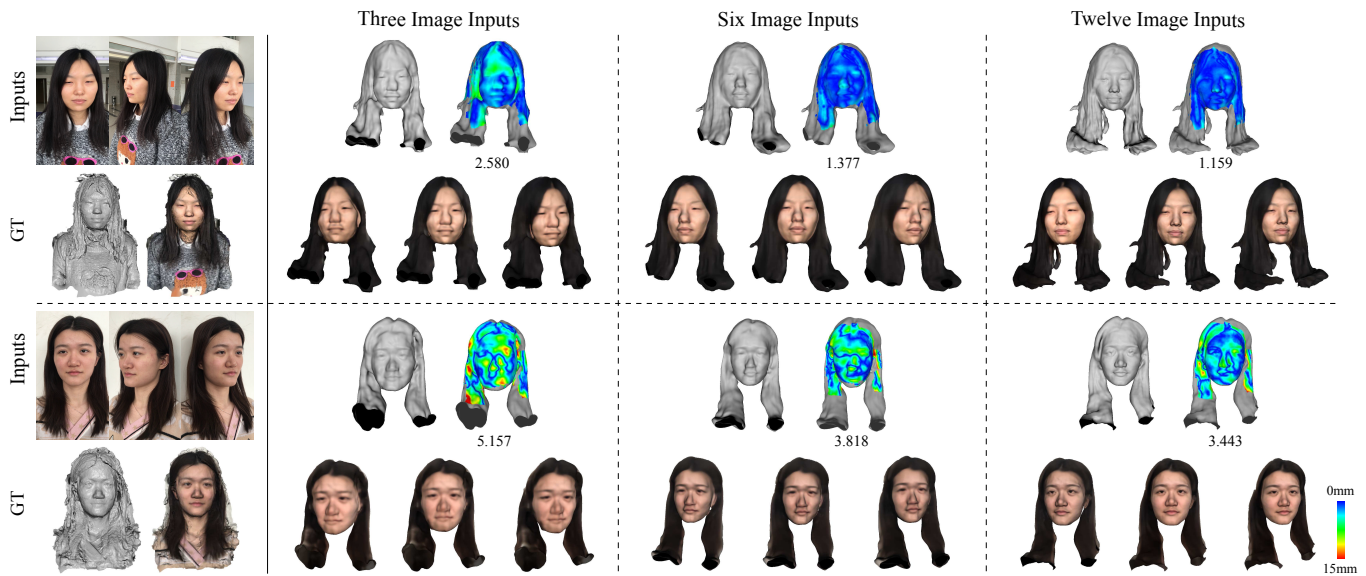}
	\caption{Reconstruction results with geometric errors from three, six and twelve input images of the phone-captured data. We take three portrait images under the front, right and left views as input for the reconstruction from three input images. Another three and nine portrait images are used for the reconstruction from six and twelve input images, respectively. Reconstructed 3D head models, their corresponding geometric errors and textured meshes are shown, respectively.}
	\label{fig:phone_multi}
\end{figure*}

\subsection{Comparisons}
\label{Sec:comparisons}

\textbf{Comparisons with multi-view stereo for general object reconstructions.} We compare our method with Colmap \cite{schonberger2016pixelwise}, OpenSfM \cite{opensfm} and IDR \cite{yariv2020multiview}. Both Colmap \cite{schonberger2016pixelwise} and OpenSfM \cite{opensfm} are traditional optimization-based multi-view stereo methods that perform the camera calibration and object reconstruction mainly based on some matched hand-crafted features. IDR \cite{yariv2020multiview} combines implicit neural representation and differentiable rendering to model the geometry and appearance of an object from multi-view images. In this paper, we use IDR as our baseline. To further improve the reconstruction accuracy and robustness, we incorporate three different head priors to guide the reconstruction processing. First, we compare these four methods on our indoor-collected data and photo-captured data with a few input images. For all of the methods, we take 12 multi-view portrait images as the input and compare the geometric accuracy of the reconstructed 3D head models qualitatively and quantitatively. In this case, the average running time of Colmap \cite{schonberger2016pixelwise}, OpenSfM \cite{opensfm}, IDR \cite{yariv2020multiview} and our method are 0.07h, 0.15h, 3.5h and 5.7h (without prior constructions) respectively, on a single GeForce RTX 2080 Ti GPU. Fig.~\ref{fig:dataset_mvs} qualitatively shows that compared with these methods, our method can recover more accurate and complete 3D head models. We also carry out a quantitative comparison of these methods on the phone-captured data and compute geometric errors by first applying a transformation with seven degrees of freedom (six for rigid transformation and one for scaling) to align the reconstructed head models with the ground truth models, and then computing the point-to-point distance to the ground truth models. Fig.~\ref{fig:phone_mvs} shows that the average geometric errors of Colmap \cite{schonberger2016pixelwise}, OpenSfM \cite{opensfm}, IDR \cite{yariv2020multiview} and our method are 1.984mm, 3.500mm, 2.669mm and 2.301mm (\emph{dense}), respectively. Due to the sparsity of Colmap's \cite{schonberger2016pixelwise} reconstruction models, and particularly the incompleteness of hair regions, Colmap \cite{schonberger2016pixelwise} has some advantages in the error calculation. For a fair comparison, we further compute a \emph{sparse} geometric error, by first finding the points on our reconstructed model corresponding to the Colmap's \cite{schonberger2016pixelwise} and then computing the point-to-point distance from these corresponding points to the ground truth model. In this case, the average geometric error of our method is 1.986mm (\emph{sparse}), which is almost the same as Colmap's average error. Meanwhile, our method can greatly improve the completeness of the reconstructed models. 

Second, we make a comparison on the publicly available dataset 3DFAW \cite{pillai20192nd} with dense input images. In this case, we take 30 multi-view portrait images as the input for all of the methods and compute the reconstruction errors as the point-to-point distances from the reconstructed models to the ground truth models. The average geometric errors of Colmap \cite{schonberger2016pixelwise}, OpenSfM \cite{opensfm}, IDR \cite{yariv2020multiview} and our method are 4.111mm, 5.613mm, 4.236mm and 3.195mm respectively, with the average running time about 0.26h, 0.40h, 5h and 10h (without prior constructions) on a single GeForce RTX 2080 Ti GPU. We show three quantitative comparison examples in Fig.~\ref{fig:public_mvs}. These two comparison experiments show that our method can improve the reconstruction accuracy and completeness with both a few inputs (12 multi-view images) and dense inputs (30 multi-view images).

\textbf{Comparisons with 3D face reconstruction from multi-view images.} To evaluate the accuracy of the face regions of our reconstructed 3D head models, we compare our prior-guided reconstruction method with some state-of-the-art deep learning-based 3D face reconstruction methods from multi-view images. Most existing methods incorporate facial prior knowledge for addressing this problem and mainly directly regress 3DMM parameters from the multi-view image inputs. These methods can recover low-frequency facial geometric structures but perform poorly in the detail reconstruction. We mainly compare our method with DFNRMVS \cite{bai2020deep} and MVFNet \cite{wu2019mvf}. For a fair comparison, we take three images under the front, left and right views as input to our method and the compared methods. The average running times of DFNRMVS \cite{bai2020deep}, MVFNet \cite{wu2019mvf} and our method are approximately 0.135min, 0.263min, and 162min (without prior constructions) respectively, on a single GeForce RTX 2080 Ti GPU. The qualitative results in Fig.~\ref{fig:dataset_mvsface} demonstrate that our method can better recover face regions with fine facial details such as wrinkles. For the quantitative evaluation, we compute a geometric error for each reconstructed model from phone-captured images by first applying a transformation to align its face region with the target face region on the ground-truth model and then computing the point-to-point distance to the target face region. The average geometric errors of DFNRMVS \cite{bai2020deep}, MVFNet \cite{wu2019mvf} and our method on the phone-captured data shown in Fig.~\ref{fig:phone_mvsface} are 2.044mm, 1.717mm and 1.228mm, respectively. The average geometric errors on the publicly available dataset 3DFAW \cite{pillai20192nd} of DFNRMVS \cite{bai2020deep}, MVFNet \cite{wu2019mvf} and our method are 1.518mm, 1.434mm and 1.310mm respectively. Fig.~\ref{fig:public_mvsface} shows the quantitative results for the three subjects. It can be seen that our method outperforms other methods due to prior guidance and flexible implicit neural representation.

\subsection{Additional Results}

With the guidance of head priors, our proposed method supports reconstructions from a few multi-view input images. We demonstrate this by implementing our method with different numbers of input multi-view portrait images. For qualitative comparison, we show the results with three and twelve input images in Fig.~\ref{fig:dataset_multi}. For quantitative results, we take three, six and twelve portrait images from phone-captured data as input and compute the point-to-point distance described in Sec.~\ref{Sec:comparisons} as their geometric errors. Fig.~\ref{fig:phone_multi} illustrates that our method can perform reconstructions from a few image inputs and better results are achieved with more input images with the average geometric error ranging from 3.869mm to 2.598mm and 2.301mm.
\section{Conclusion}

We propose a prior-guided implicit neural rendering network to reconstruct a high-quality integrated 3D head model from a few multi-view portrait images. We first extract three different head priors from multi-view image inputs, including the facial prior knowledge, head semantic segmentation information and 2D hair orientation maps. Specifically, facial prior knowledge provides an initial face geometric structure, semantic segmentation information can avoid unreasonable facial organ shapes and maintain the head geometric structure, and 2D hair orientation maps can improve the reconstruction accuracy of the hair region. Then, with the guidance of this prior knowledge, we introduce three prior-based loss terms into an implicit neural rendering network to improve the reconstruction accuracy and robustness of the 3D head model. Extensive experiments demonstrate that our method outperforms state-of-the-art multi-view stereo methods for the general object reconstruction and 3D face reconstruction methods from multi-view images. Additionally, our proposed approach can perform reconstructions from different numbers of multi-view portrait images, and reasonable results can be obtained even with very sparse inputs.

\noindent \textbf{Acknowledgement}
This research is partially supported by National Natural Science Foundation of China (No. 62122071), the Youth Innovation Promotion Association CAS (No. 2018495), ``the Fundamental Research Funds for the Central Universities'' (No. WK3470000021).

{\small
	\bibliographystyle{IEEEtran}
	\bibliography{mybibfile}
}

\begin{IEEEbiography} 
[{\includegraphics[width=1in]{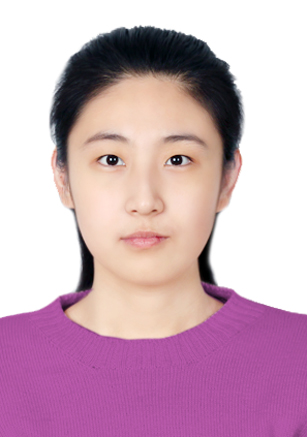}}]{Xueying Wang} is a Ph.D. student at the School of Mathematical Sciences, University of Science and Technology of China. She obtained her bachelor's degree from Northeastern University in 2017. Her research interests include Computer Vision and Computer Graphics.
\end{IEEEbiography}

\begin{IEEEbiography} 
[{\includegraphics[width=1in]{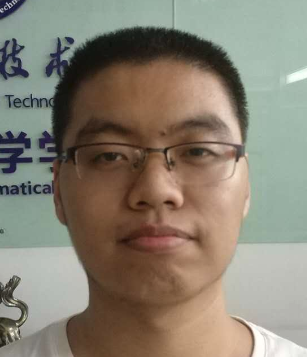}}]{Yudong Guo} is a Ph.D. student at the School of Mathematical Sciences, University of Science and Technology of China. He obtained his bachelor's degree from the same university in 2015. His research interests include Computer Vision and Computer Graphics.
\end{IEEEbiography}

\begin{IEEEbiography} 
[{\includegraphics[width=1in]{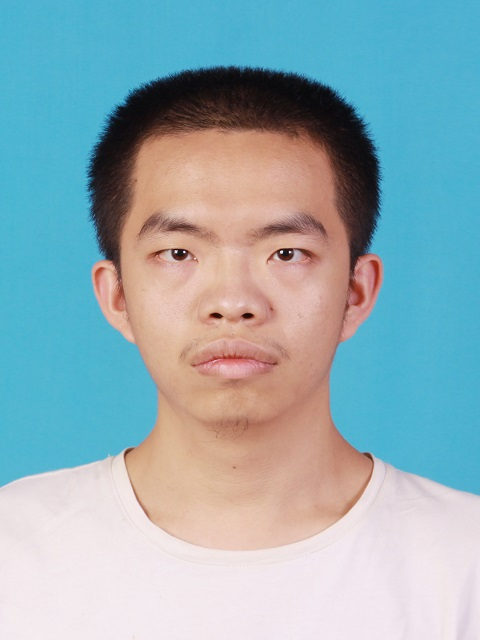}}]{Zhongqi Yang} is a Postgraduate student at the School of Mathematical Sciences, University of Science and Technology of China. He obtained his bachelor's degree from the same university in 2020. His research interests include Computer Vision and Computer Graphics.
\end{IEEEbiography}

\begin{IEEEbiography} 
[{\includegraphics[width=1in]{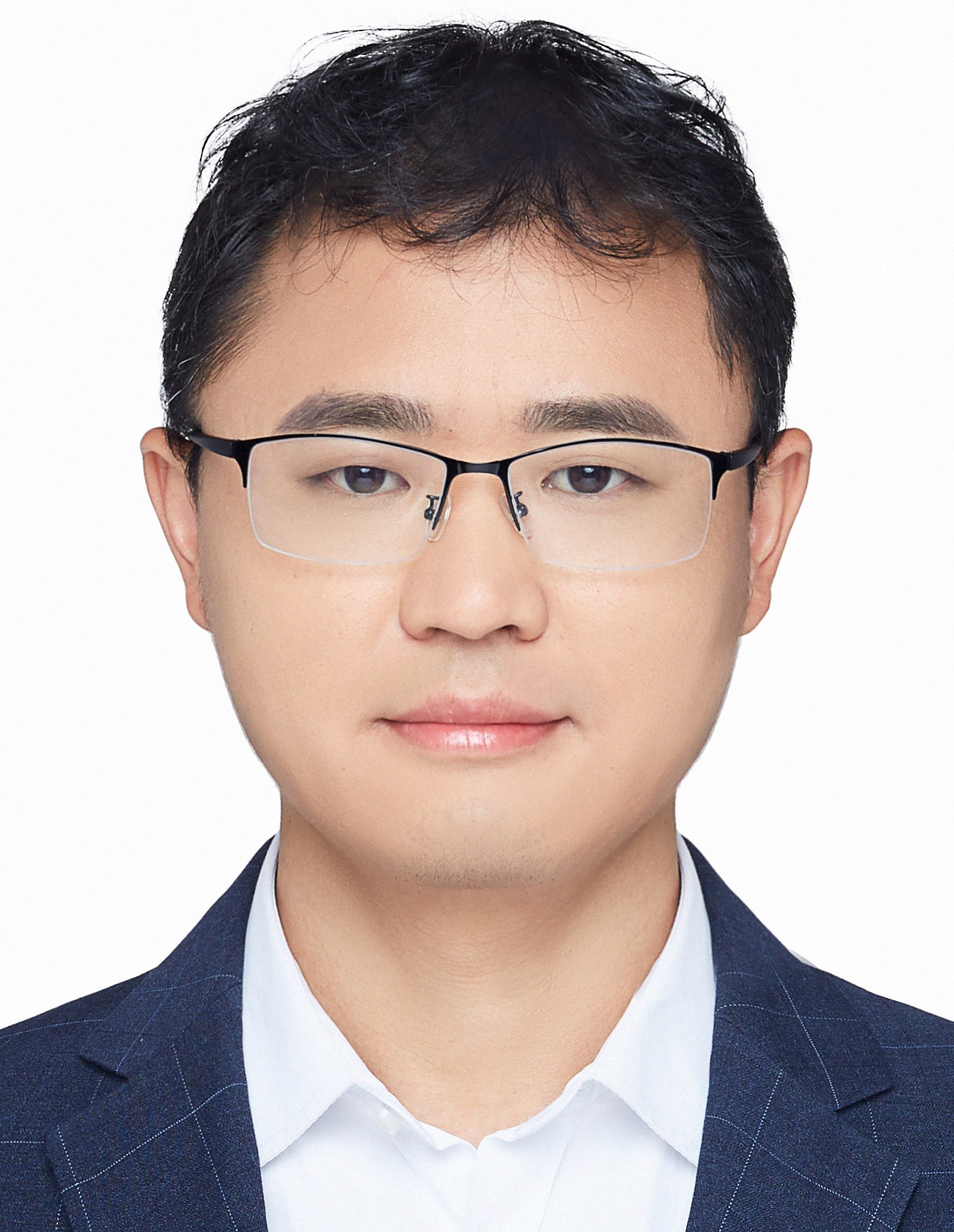}}]{Juyong Zhang} is an associate professor in the School of Mathematical Sciences at the University of Science and Technology of China. He received his BS degree from University of Science and Technology of China in 2006, and Ph.D. degree from Nanyang Technological University, Singapore. His research interests include computer graphics, 3D computer vision and numerical optimization. He is currently an Associate Editor for IEEE Transaction on Multimedia and the Visual Computer Journal. 
\end{IEEEbiography}

\end{document}